\documentclass[journal]{IEEEtran}

\ifCLASSINFOpdf
\else
  \usepackage[dvips]{graphicx}
\fi

\usepackage{bbm}
\usepackage{url}
\usepackage{color, soul}
\usepackage{cite}
\usepackage[colorlinks,linkcolor=black, citecolor=black]{hyperref}
\usepackage{multirow}
\usepackage{amsmath}
\usepackage{bm}
\usepackage{graphicx}
\usepackage{comment}
\usepackage{longtable}
\usepackage{xparse}
\usepackage{booktabs}
\usepackage{amsmath}
\usepackage{amssymb}
\usepackage{caption}
\usepackage[caption=false,font=footnotesize,labelfont=rm,textfont=rm, justification=centering]{subfig}
\usepackage{textcomp}
\usepackage{stfloats}
\usepackage{array}
\usepackage{colortbl}% 表格颜色支持
\usepackage[dvipsnames, svgnames, x11names]{xcolor}% 颜色支持\usepackage{colortbl}% 表格颜色支持
\usepackage{ulem}
\usepackage{amsmath}
\usepackage{amssymb}
\usepackage{setspace}
\usepackage{colortbl}

\begin{document}

 % \text{+}

\title{FluentEditor2: Text-based Speech Editing by Modeling Multi-Scale Acoustic and Prosody Consistency}
 
\author{Rui Liu, \IEEEmembership{Member, IEEE}, Jiatian Xi, Ziyue Jiang and Haizhou Li  \IEEEmembership{Fellow, IEEE}

% \vspace{-2mm}
\thanks{
The research by Rui Liu was funded by the Young Scientists Fund (No. 62206136) and the General Program (No. 62476146) of the National Natural Science Foundation of China,  Guangdong Provincial Key Laboratory of Human Digital Twin (No. 2022B121201 0004),
% the High-level Talents Introduction Project of Inner Mongolia University (No. 10000-22311201),
and the ``Inner Mongolia Science and Technology Achievement Transfer and Transformation Demonstration Zone, University Collaborative Innovation Base, and University Entrepreneurship Training Base'' Construction Project (Supercomputing Power Project) (No.21300-231510). 
The research by Haizhou Li was partly supported by Internal Project Fund from Shenzhen Research Institute of Big Data (Grant No.~T00120220002), and Shenzhen Science and Technology Research Fund (Fundamental Research Key Project Grant No.~JCYJ20220818103001002).
}

\thanks{Rui Liu and Jiatian Xi are with the Department of Computer Science, Inner Mongolia University, Hohhot 010021, China. 
(e-mail: liurui\_imu@163.com).}

\thanks{Ziyue Jiang is with the Zhejiang University, Hangzhou, China. (e-mail: ziyuejiang@zju.edu.cn).}

\thanks{Haizhou Li is with School of Data Science, The Chinese University of Hong Kong, Shenzhen 518172, China. He is also with University of Bremen, Faculty 3 Computer Science / Mathematics, Enrique-Schmidt-Str. 5 Cartesium, 28359 Bremen, Germany (e-mail: haizhouli@cuhk.edu.cn). }
}

\markboth{PREPRINT MANUSCRIPT OF IEEE/ACM TRANSACTIONS ON AUDIO, SPEECH, AND LANGUAGE PROCESSING}%
{Shell \MakeLowercase{\textit{et al.}}: Bare Demo of IEEEtran.cls for IEEE Journals}

\maketitle

\begin{abstract}
Text-based speech editing (TSE) allows users to edit speech by modifying the corresponding text directly without altering the original recording. Current TSE techniques often focus on minimizing discrepancies between generated speech and reference within edited regions during training to achieve fluent TSE performance. However, the generated speech in the edited region should maintain acoustic and prosodic consistency with the unedited region and the original speech at both the local and global levels. To maintain speech fluency, we propose a new fluency speech editing scheme based on our previous \textit{FluentEditor} model, termed \textit{\textbf{FluentEditor2}}, by modeling the multi-scale acoustic and prosody consistency training criterion in TSE training. Specifically, for local acoustic consistency, we propose \textit{hierarchical local acoustic smoothness constraint} to align the acoustic properties of speech frames, phonemes, and words at the boundary between the generated speech in the edited region and the speech in the unedited region. For global prosody consistency, we propose \textit{contrastive global prosody consistency constraint} to keep the speech in the edited region consistent with the prosody of the original utterance. Extensive experiments on the VCTK and LibriTTS datasets show that \textit{FluentEditor2} surpasses existing neural networks-based TSE methods, including Editspeech, Campnet, A$^3$T, FluentSpeech, and our Fluenteditor, in both subjective and objective. Ablation studies further highlight the contributions of each module to the overall effectiveness of the system. Speech demos are available at: \url{https://github.com/Ai-S2-Lab/FluentEditor2}.
\end{abstract}


% Abstract---ororor
% Text-based speech editing (TSE) allows users to modify text records instead of altering audio directly, enabling operations like cut, copy, and paste. However, current TSE techniques often focus solely on minimizing discrepancies between edited and reference speech segments, neglecting broader fluency considerations and the integration of edited and non-edited segments. Our proposed method, FluentEditor2, addresses these limitations by leveraging feature extraction for acoustic and prosody characteristics. It balances segmental smoothness and global consistency to ensure seamless transitions between edited and non-edited regions. FluentEditor2 offers delete, insert, and replace operations, and employs hierarchical techniques to maintain speech quality and coherence. It introduces an adaptive approach with contrast learning for prosody features, enhancing overall performance. Experiments on the VCTK dataset demonstrate FluentEditor2's superiority over existing TTS technologies, including Campnet and A$^3$T, and Fluentspeech, with detailed ablation experiments confirming the efficacy of each module.

\IEEEpeerreviewmaketitle

% \vspace{-3mm}
\section{Introduction}
\label{sec:intro}

% {\color{red}
% \hl{dont remove !}

% para1:
% style transfer is very important.

% para2:
% cross-speaker emotion transfer is very interesting.

% para3: 
% However, if the reference speech contains noise, the quality of synthesized speech will drop.

% para4:
% To address this issue, we propose our model, NCE-TTS.

% }

% {\color{red}
% para1:
% style transfer is very important.
% }

\IEEEPARstart{T}{ext-based speech editing} (TSE) \cite{jin2017voco} allows for modification of the audio by editing the transcript rather than the audio itself \cite{jin2017voco}. With the rapid development of the internet, audio-related media sharing has become a prevalent activity in our daily lives. Note that TSE can bring great convenience to the audio generation process and be applied to a variety of areas with personalized voice needs, including video creation for social media, games, and movie dubbing \cite{10379131,10487819}.

% 语音编辑定义
The traditional TSE approach follows the following formal definition: Given a reference audio sample $A_{\text{ref}}$ along with its corresponding textual transcript $T_{\text{ref}}$, aligned using a Montreal Forced Aligner (MFA) \cite{mcauliffe2017montreal}, the TSE task involves converting the edited text $T_{\text{edited\_text}}$ into the corresponding audio sequence $A_{\text{edited\_audio}}$. 
The objective is to ensure the final edited audio $A_{\text{edited}}$ remains imperceptibly close to the original $A_{\text{ref}}$, preserving the naturalness and fluidity of the speech. 
During inference, the resulting audio segment is combined with the unedited portion of the reference audio, resulting in a natural and fluent output \cite{jiang2023fluentspeech}.
Please note that the critical challenge for TSE lies in ensuring a fluent transition \cite{liu2023fluenteditor} at the editing boundaries between $A_{\text{ref}}$ and $A_{\text{edited}}$.
In other words, fluent transitions are essential to avoid perceptible artifacts that may distract the listener, especially in the editing region. Discontinuities at the boundary can degrade the overall audio quality.
% making it vital to apply precise trimming and concatenation techniques. 
% Therefore, achieving fluent transitions in the editing regions is paramount for maintaining high-quality audio output.
Therefore, it is crucial to achieve smooth transitions at the boundary between the editing area and the non-editing area to maintain high-quality audio output \cite{morrison2021context}.

% Subsequently, utilizing $A_{\text{ref}}$, the final target audio $A_{\text{edited}}$ is synthesized through appropriate trimming and concatenation operations. The objective of the task is to ensure a seamless similarity between $A_{\text{ref}}$ and $A_{\text{edited}}$ to the extent that the differences become imperceptible to the listener.

In recent years, many efforts have been made to build neural networks-based TSE models inspired by neural text-to-speech (TTS) models.
For example, 
% one approach involves utilizing a series of neural networks to seamlessly integrate segments from the same speaker, leveraging pitch and prosody features to achieve natural editing~\cite{morrison2021context}. 
in EditSpeech~\cite{tan2021editspeech}, the authors divided the given speech utterance into ``to-modify'' and ``non-modify'' regions according to the edited text and speech-text alignment, and generated the new ``modified'' speech frames using a duration based auto-regressive neural TTS conditioned on the ``non-modify'' frames. 
The CampNet ~\cite{wang2022campnet} conducted mask training on a context-aware neural network based on Transformer to improve the quality of the edited voice. 
A$^3$T ~\cite{bai20223} suggested an alignment-aware acoustic and text pretraining method, which can be directly applied to speech editing by reconstructing masked acoustic signals through text input and acoustic text alignment.
More recently, the diffusion model has gradually become the backbone of the neural networks-based TSE with remarkable results. For example, 
EdiTTS ~\cite{tae22_interspeech, zen19_interspeech} takes the diffusion-based TTS model as the backbone and proposes a score-based TSE methodology for fine-grained pitch and content editing.
FluentSpeech ~\cite{jiang2023fluentspeech} proposes a context-aware diffusion model that iteratively refines the modified mel-spectrogram with the guidance of context features.

However, during training, existing TSE approaches primarily rely on constraining the Euclidean Distance~\cite{reyes2016spectrum} between the predicted and ground truth mel-spectrogram to ensure the naturalness of the edited speech. While they incorporate contextual information \cite{wang2022campnet,jiang2023fluentspeech} to mitigate the over-smoothing problem, their objective functions are not specifically designed to ensure a more refined and fluent speech output~\cite{tseng2005fluent, liu2021expressive}. Therefore, it is imperative to seek out more efficient speech fluency modeling in order to achieve seamless and fluent transitions in the editing boundaries.

In this study, we identify two significant challenges that must be addressed to model speech editing fluency effectively, which are \textit{\textbf{Local Acoustic Consistency}} and \textit{\textbf{Global Prosody Consistency}}. These two challenges are called \textbf{multi-scale consistency}.

\begin{enumerate}
    \item \textit{\textbf{Local Acoustic Consistency}}: The smoothness of the transition between the edited region and its neighboring segments must resemble real concatenation points. More important, such smoothness should be expressed in the hierarchical structure \cite{glass1988multi}, which includes frames, phonemes, and words, of the acoustic signals \cite{rabiner2007introduction}. This ensures natural transitions throughout the edited speech. 

    \item \textit{\textbf{Global Prosody Consistency}}: The prosodic style of the synthesized audio in the edited region should remain consistent with the overall prosody of the original utterance \cite{honda2004physiological, qian2021global}. 
\end{enumerate}

% {\color{blue} 

To address these challenges, we propose \textit{FluentEditor2}, a novel speech editing framework that introduces new training criteria for the multi-scale acoustic and prosody consistency. Specifically, \textit{FluentEditor2} introduces two major innovations: 1) \textit{\textbf{Hierarchical Local Acoustic Smoothness Consistency Constraint}} ($\mathcal{L}_{HLAC}$), which computes the concatenation costs \cite{blouin2002concatenation} at multiple granularities (frame, phoneme, and word), ensuring that the \textit{acoustic information} at the editing boundaries closely resembles real concatenation points across these hierarchical levels. Note that the concatenation cost is inspired by the traditional unit selection TTS  \cite{hunt1996unit, wouters2000unit, blouin2002concatenation, dong2006unit, fu2018deep}. We know that the main purpose of unit selection TTS is to stitch the selected multiple acoustic units into a smooth continuous speech fragment \cite{hunt1996unit, dong2006unit, fu2018deep}. 
The concatenation cost is the key to achieving this goal, ensuring a smooth transition between different acoustic units at concatenation points or boundaries \cite{wouters2000unit, chappell2002comparison}. 
Note that to achieve the hierarchical constraint of $\mathcal{L}_{HLAC}$ at multiple granularities (frame, phoneme, and word), we adopt the new word-level masking strategy instead of the random frame masking \cite{wang2022campnet,jiang2023fluentspeech} in traditional TSE methods.
2) \textit{\textbf{Contrastive Global Prosody Consistency Constraint}} ($\mathcal{L}_{CGPC}$), which employs a contrastive learning mechanism~\cite{le2020contrastive} to ensure that the prosodic features of the edited region remain consistent with the surrounding context of the original utterance while distinguishing it from other utterances in the same batch.

Subjective and objective evaluations on the VCTK~\cite{veaux2017cstr} and LibriTTS~\cite{zen19_interspeech} datasets show that \textit{FluentEditor2} significantly outperforms the previous \textit{FluentEditor} and all state-of-the-art TSE baselines in both acoustic and prosody consistency while maintaining a fluency performance that is nearly indistinguishable from natural speech.
% }

While this work shares a similar motivation with our previous
\textit{FluentEditor} work \cite{liu2023fluenteditor} in terms of acoustic and prosody consistency modeling, it is different in many ways. 1) For local acoustic consistency, \textit{FluentEditor} just considers the frame-level consistency constraint, while overlooks the phoneme- and word-level consistency; 2) To support the \textit{Hierarchical Local Acoustic Smoothness Consistency Constraint}, \textit{FluentEditor2} adopt the new word-level masking strategy, while \textit{FluentEditor} employ the random frame masking; 3) For global prosody consistency, \textit{FluentEditor} just adopt Mean Squared Error (MSE) \cite{chapman2016statistical} loss to pull close the prosody feature between the edited region and the original speech, while \textit{FluentEditor2} propose a novel contrastive learning-based loss term to implement a stronger constraint; 4) From an experimental perspective, \textit{FluentEditor2} conduct a more comprehensive and detailed subjective and objective experiment on more benchmarking datasets, including VCTK and LibriTTS. 
In a nutshell, the main contributions of this work can be summarized as follows:
\begin{itemize}
    \item We propose a novel fluent TSE scheme, termed FluentEditor2. To the best of our knowledge, this is the first attempt to consider the multi-scale acoustic and prosody consistency training constraint for TSE task.
    \item We propose \textit{Hierarchical Local Acoustic Smoothness Consistency} ($\mathcal{L}_{HLAC}$) and \textit{Contrastive Global Prosody Consistency} ($\mathcal{L}_{CGPC}$) constraints to conduct the fluency modeling for TSE. Note that the $\mathcal{L}_{HLAC}$ takes into account multiple granularities such as frames, phonemes, and words simultaneously.
    \item Comprehensive subjective and objective experimental validate that the proposed FluentEditor2 outperforms all advanced TSE baselines in terms of naturalness and fluency.
\end{itemize}

The rest of this paper is organized as follows. Section \ref{sec:rw} reviews related work. Section \ref{sec:pm} introduces the proposed method. Section \ref{sec:exp} describes the experiments and setup in detail. Section \ref{sec:RD} presents the experimental results and discusses the performance. Finally, the paper concludes in Section \ref{sec:con}.

\section{Related Works}
\label{sec:rw}

\subsection{Speech Editing}
Traditional speech editing methods often rely on digital signal processing (DSP) techniques applied directly to audio~\cite{whittaker2004semantic, rubin2013content, baume2018contextual, descript}. To simplify user interaction, text-based speech editing (TSE) systems were developed, allowing users to insert, delete, or replace speech by modifying the text directly. However, these methods often struggle to match the prosody of the edited regions with the surrounding speech. To address this, Morrison~\cite{morrison2021context} introduced a method that predicts duration and pitch based on surrounding context and uses the TD-PSOLA algorithm~\cite{moulines1990pitch} for prosody modification. While this improves fluency, it cannot generate new words not found in the original transcript. More recently, neural TTS approaches have shown promise~\cite{sisman2020overview, mohammadi2017overview, sun2016phonetic, jin2017voco, jin2018speech}. Editspeech~\cite{tan2021editspeech} preserves coherent prosody by employing bidirectional neural models, while CampNet~\cite{wang2022campnet} uses cross-attention to better model text-audio relationships, though it suffers from slow convergence. A$^3$T~\cite{bai20223} improves speech quality with alignment-aware pretraining integrated into Conformer models~\cite{gulati20_interspeech, guo2021recent}. Diffusion models~\cite{croitoru2023diffusion, jeong21_interspeech, popov2021grad} have emerged as a powerful tool for TSE. EdiTTS~\cite{tae22_interspeech} and FluentSpeech~\cite{jiang2023fluentspeech} leverage diffusion-based techniques to refine prosody and smooth transitions in edited speech. AttentionStitch~\cite{DBLP:journals/corr/abs-2403-04804} combines pre-trained TTS models with attention mechanisms to enhance boundary stitching. 

Despite these advancements, many current approaches focus on modifying the underlying model architectures, often overlooking the importance of designing specific loss functions that directly target prosodic fluency and coherence. This work identifies two significant challenges that are Local Acoustic Consistency and Global Prosody Consistency and proposes two new Fluency-Aware Training criterions.

\subsection{Multi-Scale Acoustic Attributes}

The speech signal is structured by different basic units, from fine to coarse, which are frames, phonemes and words \cite{simpson1989lexical}. This natural structure is unique to speech and contains extensive paralinguistic information such as fluency, articulation, prolongation and rhythm \cite{schafer1975digital}.
To this end, most works are explored toward hierarchical multi-granularity learning \cite{lin2020automatic} to improve system performance. For example, in speech emotion recognition, SpeechFormer \cite{chen22_interspeech} introduced a hierarchical efficient framework that incorporates speech characteristics for better emotion recognition, leveraging multi-scale feature extraction across different linguistic levels. In speech synthesis, MsEmoTTS \cite{lei2022msemotts} also leveraged multi-scale emotional control to enhance expressiveness, enabling more precise and nuanced emotion representation.
Hono et al. \cite{hono20_interspeech} proposed a hierarchical multi-grained generative model to enhance expressiveness. Li et al. \cite{li21r_interspeech} focused on multi-scale style control for improved synthesis. Ren et al. \cite{ren2021portaspeech} introduced PortaSpeech, a lightweight, high-quality generative model. It proposed a linguistic encoder with mixture alignment combining hard inter-word alignment and soft intra-word alignment, which explicitly extracts Word-to-Phoneme semantic information. Lei et al. \cite{lei2023msstyletts} developed MSStyleTTS for enhancing coherence using multi-scale style features. More recently, Jiang et al. \cite{jiang24d_interspeech} investigated hierarchical prosody modeling for zero-shot synthesis, improving nuanced prosodic control.

Inspired by the above works, we consider the hierarchical acoustic smoothing consistency in the TSE task. We believe that speech frames, phonemes, and words at the splice between edited and unedited speech should exhibit acoustic smoothness, and such hierarchical speech modeling is helpful for better learning speech fluency.

\subsection{Contrastive Learning}
\subsubsection{Basic knowledge of contrastive learning}
% The goal of contrastive learning is to train an encoder to produce similar representations for similar data instances while ensuring that representations of dissimilar instances are as distinct as possible.
Contrastive learning \cite{hadsell2006dimensionality, chen2020simple, viana2023multi} aims to train an encoder to produce similar representations for data instances that are alike while ensuring that representations for dissimilar instances remain as distinct as possible \cite{hadsell2006dimensionality}.  This framework is grounded in the idea that by learning to differentiate between similar and dissimilar data, the encoder can capture meaningful patterns in the data space \cite{hadsell2006dimensionality}.

Specifically, Chen et al. ~\cite{chen2020simple} suggested that training the relative relationships within a sample space—specifically, contrastive relationships between pairs of data—is sufficient to represent the vectors in that space. 
The contrastive loss function encourages the model to push similar samples closer and dissimilar samples farther apart. The loss function is defined as:

\begin{equation}
\begin{aligned}
\ell_{i,j} = -\log \frac{\exp(\text{sim}(z_i, z_j)/\tau)}{\sum_{k=1}^{2N} {\mathbbm{1}}_{[k \neq i]} \exp(\text{sim}(z_i, z_k)/\tau)}
\end{aligned} 
\end{equation}

where $\text{sim}(z_i, z_j)$ represents the cosine similarity between two latent vectors $z_i$ and $z_j$, and $\tau$ is a temperature parameter controlling the sharpness of the distribution. The indicator function $\mathbbm{1}_{[k \neq i]}$ ensures that only distinct pairs are considered, excluding self-pairs. 

In summary, contrastive learning provides a framework to train encoders to learn discriminative representations by maximizing the similarity between similar data instances and minimizing the similarity between dissimilar ones. This is achieved through the optimization of a contrastive loss function, which drives the model to capture the underlying structure of the data space.
    
\subsubsection{Applications of contrastive learning in speech tasks}

Contrastive learning has achieved considerable success in computer vision, particularly with the introduction of the SimCLR framework~\cite{chen2020simple}, which learns visual representations by maximizing the similarity between positive pairs. SimCLRv2~\cite{chen2021exploring} further improves performance by employing advanced data augmentation techniques. In the field of speech processing, contrastive learning is still an emerging area of research, but it has shown promising results.    Early works such as~\cite{yang2019reducing, bose-etal-2018-adversarial, kharitonov2021data, koepke2022audio} have demonstrated that contrastive learning can be effective in tasks like speech representation learning, adversarial training, and data augmentation for speech models.
More recent advancements have focused on learning style representations from text data. For instance,\cite{wu22e_interspeech} explores using contrastive learning to derive style representations from large text corpora.    Similarly, CALM\cite{meng22c_interspeech} leverages contrastive learning to optimize the correlation between stylistic text features, enhancing the generation of stylistically consistent speech. Furthermore, DCTTS~\cite{wu2024dctts} introduces a discrete diffusion model with contrastive learning to improve alignment between text and speech, which in turn enhances sampling rates and the overall quality of synthesized speech.

In this work, we introduce contrastive learning into the TSE task, which forces the prosodic features of the edited regional speech to be close to the global prosody of the original speech and far from other samples.
% Contrastive learning has been successfully applied in computer vision, notably through the SimCLR framework~\cite{chen2020simple}, which learns visual representations by maximizing the similarity between positive pairs. SimCLRv2~\cite{chen2021exploring} builds upon this by introducing advanced data augmentation techniques to enhance the benefits of pertaining. Furthermore, the application of contrastive learning in the field of speech ~\cite{yang2019reducing, bose-etal-2018-adversarial, kharitonov2021data, koepke2022audio} is relatively nascent and requires further research. Recently, employing contrastive learning strategies to learn style representations from a vast amount of pure text has made significant strides~\cite{wu22e_interspeech}. CALM~\cite{meng22c_interspeech} optimizes the correlation between stylistic text features through contrastive learning. DCTTS~\cite{wu2024dctts} proposes a discrete diffusion model with contrastive learning to enhance alignment between text and speech and improve sampling rates.

\begin{figure*}[!t]
\centering
\setlength{\abovecaptionskip}{-0mm}   %调整图片标题与图距离
\centerline{\includegraphics[width=0.98\linewidth]{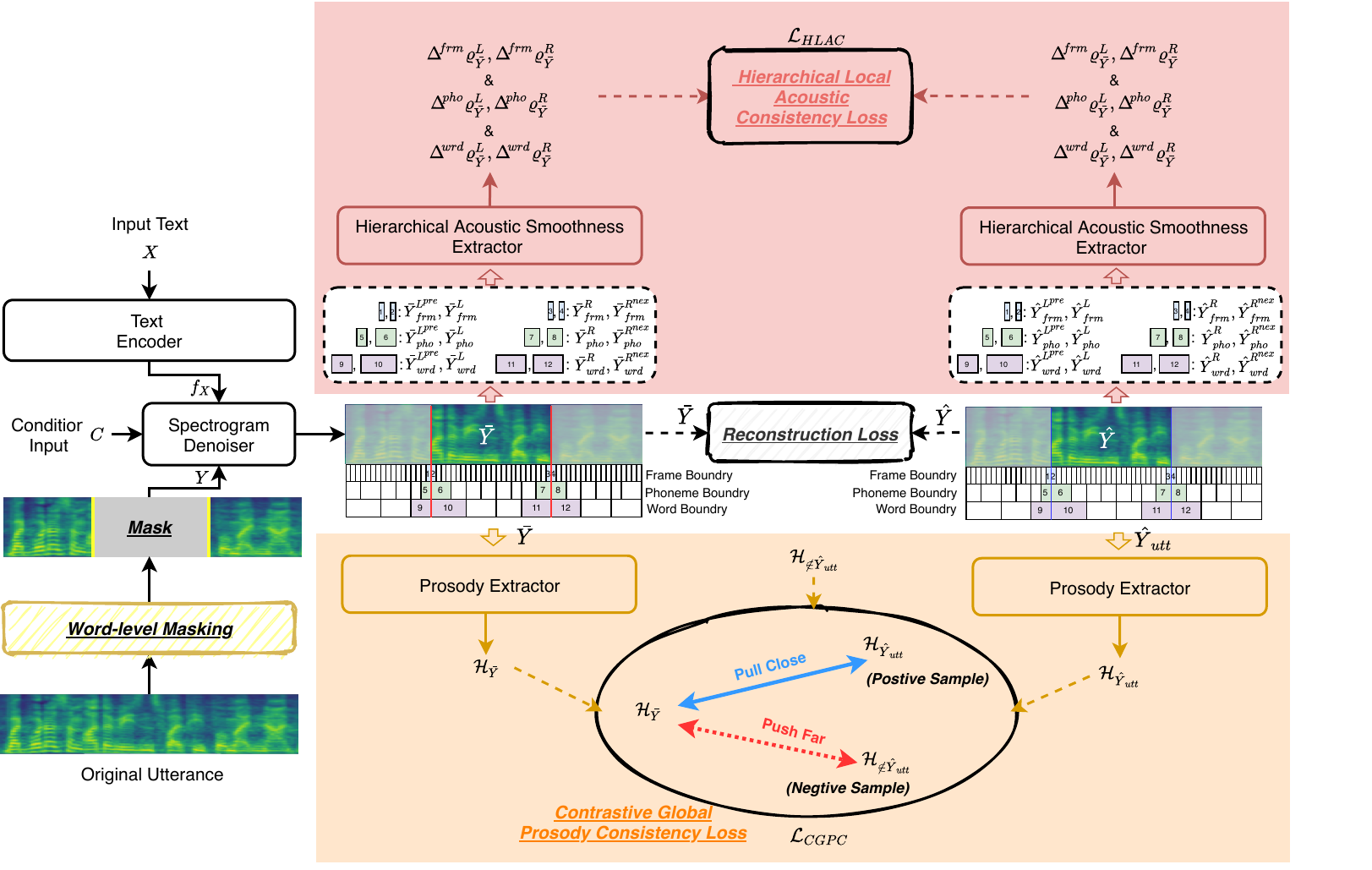}}
% \vspace{-2mm}
\caption{The overall workflow of FluentEditor2. The total loss function comprises Reconstruction Loss, and Local Hierarchical Acoustic Smoothness and Contrastive Global Prosody Consistency Losses.}
\vspace{-3mm}
\label{fig:fig1}
\end{figure*}

\section{FluentEditor2: Methodology}
\label{sec:pm}
 
\subsection{Overview Architecture}
As shown in Fig.\ref{fig:fig1}, our FluentEditor2 adopts the mask prediction-based diffusion network as the backbone, which consists of a text encoder, and a spectrogram denoiser. The spectrogram denoiser seeks to adopt the Denoising diffusion probabilistic model (DDPM) to learn a data distribution $p(\cdot)$ by gradually denoising a normally distributed variable through the reverse process of a fixed Markov Chain of length $T$.
% 扩散模型input有一些问题。
Assume that the input phoneme sequence of the input phoneme sequence is $X = (X_1, \ldots, X_{|X|})$, from which the text encoder extracts the frame-level text embedding $f_X = (f_{X_1}, f_{X_2}, \ldots, f_{X_{|X|}})$. The masked acoustic feature sequence $Y = Mask(f_X, \lambda)$ is generated by replacing entire word-level spans of $f_X$ with random vectors based on a probability $\lambda$.
The spectrogram denoiser then aggregates several inputs: the text embedding $f_X$, the masked acoustic feature $Y$, and the condition input $C$ to guide the reverse process of the diffusion model $\Theta(Y_t | t, C)$, where $t \in T$, and $Y_t$ is a noisy version of the clean input $\hat{Y}_{utt}$. 
Similar to \cite{jiang2023fluentspeech}, the condition input $C$ includes the frame-level text embedding $f_X$, the acoustic feature sequence $\hat{Y}_{utt}$, the masked acoustic feature sequence $Y$, the speaker embedding $e_{spk}$, and the pitch embedding $e_{pitch}$. 
In the generator-based diffusion models, $p_{\theta}(\hat{Y}_{utt}|Y_{t})$ is the implicit distribution imposed by the neural network $f_{\theta}(Y_t, t)$ that outputs $\bar{Y}$ given $Y_{t}$. And then $Y_{t-1}$ is sampled using the posterior distribution $q(Y_{t-1} | Y_{t}, \bar{Y})$ given $Y_{t}$ and the predicted $\bar{Y}$.

To model speech fluency, we design \textit{Hierarchical Local Acoustic Smoothness Consistency loss} $\mathcal{L}_{HLAC}$ and \textit{Contrastive Global Prosody Consistency Loss} $\mathcal{L}_{CGPC}$ separately on the basis of the original \textit{reconstruction loss}, to ensure that the acoustic and prosody performance of speech generated in the editing area is consistent with the context and the original utterance. For reconstruction loss,  we follow \cite{jiang2023fluentspeech} and employ Mean Absolute Error  (MAE) and the Structural Similarity Index (SSIM) \cite{ren2022revisiting} losses to calculate the difference between $\bar{Y}$ and the corresponding ground truth segment $\hat{Y}$.

\textit{Hierarchical Local Acoustic Smoothness Consistency loss} $\mathcal{L}_{HLAC}$ and \textit{Contrastive Global Prosody Consistency Loss} $\mathcal{L}_{CGPC}$ serve the purposes of ensuring the smoothness of connecting points between edited speech regions and neighboring acoustic segments, as well as ensuring that the local prosody of the masked speech aligns with the global prosody of the ground truth. 

In the following subsection, we will first introduce the key operations, the word-level masking strategy, for implementing $\mathcal{L}_{HLAC}$ and dive deeper into $\mathcal{L}_{HLAC}$ and $\mathcal{L}_{CGPC}$ in detail, building upon the underlying training principles of acoustic and prosody consistency losses.

\subsection{Word-level Masking Strategy}
Traditional speech editing systems often adopt random frame-level masking strategies \cite{wang2022campnet,jiang2023fluentspeech} to mask speech segments and reconstruct them during training. 
However, the random frame-level masking often disrupts phoneme or word boundaries, leading to unnatural transitions and misalignment. In addition, to support the calculation of multi-granularity acoustic smoothness at the boundary between edited and non-edited speech segments, we propose a \textit{Word-level Masking (WLM) Strategy}.
Specifically, we first obtain the ``speech frame-phoneme- word'' alignment with the help of the MFA (More details please refer to Section \ref{setup}.), and then mask all the speech frames corresponding to several consecutive words according to the word timestamp. 

In this way, the Hierarchical structure integrity of the mask regions area and the non-mask regions can be maintained, and more natural speech editing can be achieved.

% The advantage of the proposed word-level masking strategy is twofold: (1) By masking entire words, it preserves the acoustic and prosodic continuity of the utterance, which is crucial for alignment-sensitive operations such as concatenation.  This ensures smoother transitions in the edited speech.  (2) The method aligns with the natural linguistic structure of speech, where words, rather than individual frames, form the primary units of communication.  Consequently, the model can better capture contextual relationships, leading to more coherent and contextually consistent speech outputs.

\subsection{Fluency-Aware Training Criterion}

The FluentEditor2 model introduces novel loss functions aimed at improving the fluency of edited speech by ensuring both Hierarchical Local Acoustic Smoothness and Contrastive Global Prosody Consistency. These two complementary mechanisms ensure fluent transitions between edited and unedited regions, even in cases involving substantial modifications. By addressing acoustic smoothness at multiple hierarchical levels (frame, phoneme, and word), while maintaining global prosodic coherence, the model produces more fluent and coherent speech. This process is key to fluency-aware training, where both aspects are jointly optimized.

\subsubsection{\textbf{Hierarchical Local Acoustic Smoothness Consistency Loss}}

Unit-selection TTS systems traditionally rely on two critical loss functions: \textit{Target Loss} and \textit{Concatenation Loss} \cite{dong2006unit, fu2018deep, hunt1996unit}. The Target Loss evaluates the proximity of candidate units to the target \cite{fu2018deep}, while the Concatenation Loss \cite{hunt1996unit} measures the smoothness between concatenated speech units.

Building upon the principles of \textit{Concatenation Loss} \cite{hunt1996unit}, we propose the \textit{Hierarchical Local Acoustic Smoothness Consistency Loss} ($\mathcal{L}_{HLAC}$). 
This loss ensures smooth transitions across multiple hierarchical levels: frame, phoneme, and word, between the edited regions and their surrounding context. Thus, the $\mathcal{L}_{HLAC}$ loss is defined as an integration of frame-level, phoneme-level, and word-level smoothness:

\begin{equation}
% \begin{aligned}
\mathcal{L}_{HLAC}=\mathcal{L}_{AC(frm)}  +\mathcal{L}_{AC(pho)} +\mathcal{L}_{AC(wrd)}
% \end{aligned} 
% \vspace{-2mm}
\end{equation}

% The symbols $\theta_{frm}$, $\theta_{pho}$, and $\theta_{wrd}$ 
% % capture relevant mel-spectrogram acoustic information at these different hierarchical levels.
% {\color{blue} captures relevant mel-spectrogram acoustic information at different hierarchical levels, rather than traditional spectral or energetic information.}

% The $\mathcal{L}_{HLAC}$ loss constrains smoothness at both the left and right boundaries of the predicted acoustic feature $\bar{Y}$, respectively, across the three hierarchical levels. 
Take frame-level as a example, $\mathcal{L}_{AC(frm)}$ consists of two components, with $L$ and $R$ denoting the left and right boundaries, respectively:

\begin{equation}
\begin{aligned}
\mathcal{L}_{AC(frm)} = \mathcal{L}^L_{AC(frm)} + \mathcal{L}^R_{AC(frm)}
\end{aligned} 
% \vspace{-2mm}
\end{equation}

Next, we employ the \textit{Mean Squared Error} (MSE) \cite{mandhare2019generalized} to quantify the proximity between the predicted $\bar{Y}$ and the ground truth $\hat{Y}$ in terms of the Euclidean distance: 

\begin{equation} \begin{aligned}
\mathcal{L}^L_{AC(frm)} = \text{MSE}(\Delta^{frm} \varrho_{\bar{Y}}^L, \Delta^{frm} \varrho_{\hat{Y}}^L) 
\\ \mathcal{L}^R_{AC(frm)} = \text{MSE}(\Delta^{frm} \varrho_{\bar{Y}}^R, \Delta^{frm} \varrho_{\hat{Y}}^R)
\end{aligned}  
\end{equation}

Note that $\Delta \varrho_{\bar{Y}}^L$ and $\Delta \varrho_{\bar{Y}}^R$ represent the Euclidean distances between adjacent frames, extracted by the Hierarchical Acoustic Smoothness Extractor. 
For example, illustrated in Fig.\ref{fig:fig1}, the blue boxes (1-4) represent frame-level units. $\Delta^{frm} \varrho_{\bar{Y}}^L$ represents the difference between the Mel-spectrogram features at the left boundary $\hat{Y}_{\text{frm}}^{L}$ (box 2) and those in the preceding frame $\hat{Y}_{\text{frm}}^{L^{\text{pre}}}$ (box 1).
A smaller difference in these Euclidean distance values indicates smoother transitions \cite{hashim2021mobile}.
% $\Delta^{frm} \varrho_{\bar{Y}}^R$ denotes the difference between the Mel-spectrogram features at the right boundary $\bar{Y}_{\text{frm}}^{R}$ and those in the following frame $\bar{Y}_{\text{frm}}^{R^{\text{nex}}}$. 

\begin{equation} 
\begin{aligned}
\Delta^{frm} \varrho_{\bar{Y}}^L = \left| \theta(\bar{Y}_{\text{frm}}^{L}) - \theta(\bar{Y}_{\text{frm}}^{L^{\text{pre}}}) \right|
\end{aligned}  
\end{equation}

Similarly, $\Delta^{frm} \varrho_{\bar{Y}}^R$ represents the difference between the Mel-spectrogram features at the right boundary $\hat{Y}_{\text{frm}}^{R}$ (box 3) and those in the following frame $\hat{Y}_{\text{frm}}^{R^{\text{nex}}}$ (box 4).
Smoothness constraints are applied at the right boundary using $\bar{Y}_{\text{frm}}^{R}$ (box3) rom the masked region and $\bar{Y}{\text{frm}}^{R^{\text{nex}}}$ (box4) from the non-masked region, which is the frame immediately following the edited region. This ensures coherence on both sides of the edited segment.

\begin{equation} 
\begin{aligned}
\Delta^{frm} \varrho_{\bar{Y}}^R = \left| \theta(\bar{Y}_{\text{frm}}^{R^{\text{nex}}}) - \theta(\bar{Y}_{\text{frm}}^{R}) \right|
\end{aligned}  
\end{equation}

Please note that $\theta$ for frame-level means the Euclidean distance is directly calculated using the Mel-spectrogram acoustic features from adjacent frames. For phoneme and word levels, $\theta$ represents a function to compute the average value across the multiple frames that make up a phoneme or word. This ensures smoothness is evaluated based on aggregated frame information for these levels.

For phoneme-level and word-level consistency loss functions  $\mathcal{L}_{AC(pho)}$ and $\mathcal{L}_{AC(wrd)}$, we similarly apply smoothness constraints between phoneme or word boundaries to ensure smooth transitions across consecutive phonemes or words respectively. 
In Fig.\ref{fig:fig1}, the green boxes (5-8) indicate phoneme-level units, while the purple boxes (9-12) represent word-level units, showing the application of smoothness constraints at each level.
The smoothness loss functions at both levels are defined as:

\begin{equation} 
\begin{aligned}
\mathcal{L}_{AC(pho)} = \mathcal{L}^L_{AC(pho)} + \mathcal{L}^R_{AC(pho)}
\\
\mathcal{L}_{AC(wrd)} = \mathcal{L}^L_{AC(wrd)} + \mathcal{L}^R_{AC(wrd)}
\end{aligned}  
\end{equation}

% Finally, we extract high-level acoustic representations (mean, variance, max, min, and Euclidean distance) \cite{lin2021chunk} to feed into the \textit{Hierarchical Acoustic Smoothness Extractor}, enhancing the model’s ability to capture hierarchical consistency.

% {\color{blue} Finally, high-level acoustic representations (such as mean, variance, max, min, and Euclidean distance) \cite{lin2021chunk} are extracted and fed into the \textit{Hierarchical Smoothness Extractor} to enhance the model’s ability to capture consistency across the hierarchical levels.
% }

\subsubsection{\textbf{Contrastive Global Prosody Consistency Loss}}

The Contrastive Global Prosody Consistency Loss ($\mathcal{L}_{CGPC}$) ensures that the prosody features in the predicted masked region $\bar{Y}$ align with those of the original speech $\hat{Y}$. We achieve this by comparing high-level prosody representations $\mathcal{H}_{\bar{Y}}$ and $\mathcal{H}_{\hat{Y}}$, which are obtained through a pre-trained prosody extractor based on the \textit{Global Style Token (GST)} model \cite{wang2018style}.

% {\color{blue}
To enhance the model's ability to predict masked regions while preserving global prosody coherence, a \textit{contrastive learning mechanism} is integrated into the $\mathcal{L}_{CGPC}$ loss. This mechanism pulls the prosody of the masked region closer to that of the full utterance (positive sample) and pushes it away from other utterances in the batch (negative samples). The pull-close operation ensures that the masked region aligns with the overall prosody of the current utterance, while the push-far operation enforces distinction from unrelated utterances, ensuring both fluency and coherence.

For each batch $\mathcal{B}$ with $b$ samples, where $i$ refers to the index of the current utterance, positive samples $\mathcal{H}_{\hat{Y}_{utt_i}}$ are defined as the prosody features of the entire utterance. Negative samples $\mathcal{H}_{\notin \hat{Y}_{utt_i}}$ represent prosodic features from other utterances in the batch. The $\mathcal{L}_{CGPC}$ loss loss is formulated as follows:

\begin{equation}
\mathcal{L}_{\text{GCPC}} = \sum_{i=1}^{b}-\log \frac{\exp\left( \text{sim}(\mathcal{H}_{\bar{Y}}, \mathcal{H}_{\hat{Y}_{utt_i}}) / \tau \right)}
{\begin{aligned}
\exp\left( \text{sim}(\mathcal{H}_{\bar{Y}}, \mathcal{H}_{\hat{Y}_{utt_i}}) / \tau \right) \\
+ \sum_{k \neq i} \exp\left( \text{sim}(\mathcal{H}_{\bar{Y}}, \mathcal{H}_{\notin \hat{Y}_{utt_k}}) / \tau \right)
\end{aligned}}
\end{equation}

Here, $\text{sim}(\cdot, \cdot)$ represents the cosine similarity between the prosody features of the masked region $\mathcal{H}_{\hat{Y}_{utt_i}}$ and the prosody features of both the entire utterance (positive sample) and other unrelated utterances (negative samples), while $\tau$ is the temperature parameter that controls the sharpness of the contrast between positive and negative samples. The similarity is computed as:

\begin{equation}
\text{sim}(\mathcal{H}_{\bar{Y}}, \mathcal{H}_{\hat{Y}_{utt_i}}) = \frac{\mathcal{H}_{\bar{Y}} \cdot \mathcal{H}_{\hat{Y}_{utt_i}}}{\|\mathcal{H}_{\bar{Y}}\| \|\mathcal{H}_{\hat{Y}_{utt_i}}\|}
\end{equation}

Here, $c_{i j}$ denotes the similarity between prosody features of the masked region $v_{i}^{|M|}$ and other samples $v_j$, while $\tau$ is the temperature parameter used in contrastive learning.

% }

In conclusion, by combining \textit{Local Hierarchical Acoustic Smoothness Loss} and \textit{Contrastive Global Prosody Consistency Loss}, FluentEditor2 ensures that edited speech retains natural fluency and consistent prosody across masked regions.

\subsection{Run-time Inference}

In run-time, users can edit the speech output by simply modifying the corresponding text. This process allows for intuitive control of the speech through text-based editing. Specifically, users can manually define the desired modification operation, such as \textit{insertion}, \textit{replacement}, or \textit{deletion}. For each operation,  the corresponding speech segment in the text is treated as the masked region. 
% To evaluate the model's performance, we selected 50 utterances per dataset (20 insertions, 20 replacements, and 10 deletions) for subjective and generalization experiments.  The texts were edited using ChatGPT, with carefully designed prompts to ensure realistic and random modifications, allowing for an assessment of the model's ability to handle various text-based edits.
Following the approach in \cite{jiang2023fluentspeech}, our FluentEditor2 model reads both the edited text and the remaining acoustic features, $\hat{Y} - \hat{Y}_{mask}$, from the original speech. The model then predicts the acoustic feature sequence $\bar{Y}$ for the edited word or segment. 
Once $\bar{Y}$ is predicted, it is smoothly concatenated with the surrounding acoustic features $\bar{Y} - \bar{Y}_{mask}$, ensuring fluent transitions between the edited and unedited portions of the speech. This process results in the final output speech that maintains natural continuity in terms of both timing and prosody, preserving the overall speech quality while reflecting the textual edits.

\section{Experiments}
\label{sec:exp}

\subsection{Experimental Corpora}
% We validate the FluentEditor on the VCTK \cite{veaux2017cstr} dataset, which is an English speech corpus uttered by 110 English speakers with various accents. Each recording is sampled at 22050 Hz with 16-bit quantization. The precise forced alignment is achieved through Montreal Forced Aligner (MFA) \cite{mcauliffe2017montreal}. We partition the dataset into training, validation, and testing sets, randomly with 98\%, 1\%, and 1\%, respectively.

We evaluate the performance of FluentEditor2 on the VCTK \cite{veaux2017cstr} and LibriTTS \cite{zen19_interspeech} datasets. The VCTK dataset includes speech data from 110 English speakers, featuring various accents, sampled at 22050 Hz with 16-bit quantization. In contrast, the LibriTTS dataset comprises high-quality English speech recordings at a 24 kHz sampling rate from 2,456 speakers, derived from audiobooks, with a total duration of approximately 585 hours.

To ensure precise alignment between text and speech data, we employ the Montreal Forced Aligner (MFA) \cite{mcauliffe2017montreal} for accurate forced alignment of both datasets. Additionally, each dataset is partitioned into training, validation, and testing sets with proportions of 98\%, 1\%, and 1\%, respectively.

\subsection{Participant Information}
To ensure the reliability of our subjective evaluations, we conducted listening tests with participants who were native Mandarin speakers with verified English proficiency (CET-4 and CET-6). All evaluators are graduate students specializing in speech processing,  who possess solid expertise in assessing speech quality and fluency. This specialized background ensures the authority and robustness of the evaluation results.

\subsection{Experimental Setup}
\label{setup}
The configurations of text encoder and spectrogram denoiser are referred to \cite{jiang2023fluentspeech}. The diffusion steps $T$ of the FluentEditor2 system are set to 8.
% Within the Prosody Consistency Loss module, we employ GST  \cite{wang2018style} to extract prosody feature information in a 256-dimensional space.  
Following GST \cite{wang2018style}, the prosody extractor comprises a convolutional stack and an RNN. The dimension of the output prosody feature of the GST-based prosody extractor is 256.
% {\color{blue} 
To preserve the integrity of both phonemes and words, we adopt a word-level continuous masking strategy. Through extensive experimentation, we found that a masking rate of 80\% yields the best results. Section \ref{mask} will report the results for various masking ratios.
% Moreover, since \cite{bai20223} reported that the optimal masking rate for frame-level random selection was 80\%, we conducted tests around the 80\% masking rate for word-level masking as well. After varying the masking rate slightly above and below 80\%, we confirmed that 80\% remains the most effective masking rate for word-level masking.
% }
The pre-trained HiFiGAN \cite{kong2020hifi} vocoder is used to synthesize the speech waveform.
We set the batch size is 16. The initial learning rate is set at $2 \times 10^{-4}$, and the Adam optimizer \cite{DBLP:journals/corr/KingmaB14} is utilized to optimize the network.
The FluentEditor2 model is trained with 2 million training steps on one A100 GPU.
The masking ratio for WLM in FluentEditor2 is set to 80\%.

% {\color{blue} 

To obtain frame, phoneme, and word-level alignment, we follow a forced alignment process based on the MFA  \cite{mcauliffe2017montreal}, which generates time-aligned phoneme boundaries from the provided phonetic transcription and speech signals. 1) The alignment process begins with frame-level mapping, where each audio sample is converted into Mel-spectrogram frames based on the sampling rate and hop size. This establishes a temporal representation of the audio at a fine-grained level, serving as the foundation for subsequent phoneme and word-level alignments.
2) Phoneme-level alignment is derived from the time intervals provided by the MFA \cite{mcauliffe2017montreal} in the form of TextGrid files. These intervals are mapped to the corresponding Mel-spectrogram frames, ensuring accurate timing for each phoneme. Silent phonemes are accounted for to maintain alignment precision, and phoneme durations are calculated by counting the frames assigned to each phoneme.
3) The word-level alignment is achieved by extending the phoneme-level alignment using a phoneme-to-word mapping. Each frame, already linked to a phoneme, is further mapped to the corresponding word. Word durations are then computed by aggregating the frames associated with the phonemes that constitute each word.

% This hierarchical alignment framework ensures consistency across frames, phonemes, and words, enabling precise control over speech data for TSE tasks. The preservation of both frame-level and word-level alignment is critical to maintaining fluency and accuracy.
% }

\subsection{Evaluation Metrics}
% For subjective evaluation, We conduct a Mean Opinion Score (MOS) \cite{loizou2011speech} listening evaluation in terms of speech fluency, termed \textit{FMOS}. Note that FMOS allows the listener to feel whether the edited segments of the edited speech are fluent compared to the context. We keep the text content and text modifications consistent among different models to exclude other interference factors, only examining speech fluency.

Following FluentSpeech \cite{jiang2023fluentspeech}, we designed rich subjective and objective metrics, that will be described below.

For subjective evaluation, we conduct a Mean Opinion Score (MOS) listening test \cite{loizou2011speech} to assess speech quality, Fluency-aware MOS (FMOS) \cite{liu2023fluenteditor}  to evaluate fluency, and Intelligibility-aware MOS (IMOS) \cite{chen2024diffeditor} to measure intelligibility. 
Note that FMOS allows the listener to feel whether the edited segments of the edited speech are fluent compared to the context. We keep the text content and text modifications consistent among different models to exclude other interference factors, only examining speech fluency. Additionally, Comparative FMOS (C-FMOS) \cite{loizou2011speech} is employed for the ablation study to assess the impact of different modifications on fluency. 
To minimize the vocoder's impact on subjective metrics, we follow \cite{chen2024diffeditor} and insert the predicted target regions back into their original positions within the speech.

% We compute objective metrics for the entire speech to reflect overall speech information, including information within the editing regions and at the editing boundaries. Specifically, FluentEditor2 synthesizes only mel-spectrogram features and manually performs editing operations, which are then converted into corresponding speech waveforms using a vocoder. To mitigate the vocoder's influence, when calculating the objective metrics, we paste the predicted target regions from FluentEditor2 into their respective positions in the original speech. This objective comparison ensures consistency of unedited speech across all systems. Additionally, each system's test set is identical, thereby ensuring that metrics are relevant only to the speech within editing regions, ensuring accuracy and fairness in the comparative experiments. Thus, we utilize MCD \cite{kubichek1993mel}, STOI \cite{taal2010short}, and PESQ \cite{rix2001perceptual} to measure the overall quality of the edited speech.

For objective evaluation, we directly assess the WLM-reconstructed mel-spectrograms, including both the editing regions and the boundaries.  The Mel-Cepstral Distortion (MCD)~\cite{kubichek1993mel}, Short-Time Objective Intelligibility (STOI) \cite{taal2010short}, and Perceptual Evaluation of Speech Quality (PESQ)~\cite{rix2001perceptual} metrics are computed for these reconstructed spectrograms, providing objective measures of speech quality and intelligibility.  This approach ensures a rigorous and consistent comparison of the WLM reconstruction across different models, with the metrics focusing solely on the quality of the reconstructed features.
To maintain fairness and accuracy in our comparative experiments, we use an identical test set for all systems, ensuring that the metrics reflect speech quality within the editing regions.

\begin{table*}[t]
    \centering
    \setstretch{0.88}
    \caption{Objective evaluation results of comparative study.
    % * means the value achieves suboptimal.
    }
    \renewcommand\arraystretch{1.2}
    \begin{tabular}{p{3.7cm}<{\centering}|p{1.6cm}<{\centering}p{1.6cm}<{\centering}p{1.6cm}<{\centering}|p{1.6cm}<{\centering}p{1.6cm}<{\centering}p{1.6cm}<{\centering}}
        \toprule
        \multirow{2}{*}{\textbf{Method}} & \multicolumn{3}{c|}{\textbf{VCTK}} &  \multicolumn{3}{c}{\textbf{LibriTTS}} \\
         & \textbf{MCD} $(\downarrow)$   & \textbf{STOI} $(\uparrow)$  & \textbf{PESQ} $(\uparrow)$  & \textbf{MCD} $(\downarrow)$   & \textbf{STOI} $(\uparrow)$  & \textbf{PESQ} $(\uparrow)$  \\
        \midrule
        % Ground Truth & NA & NA & NA & NA  & NA & NA \\ \hline
        EditSpeech & 7.019& 0.690 & 1.426 & 5.559 & 0.663 & 1.333 \\
        CampNet & 8.693 & 0.434 & 1.340 & 7.736 & 0.294 & 1.249 \\
        A$^3$T & 6.344 & 0.750 & 1.560 & 5.447 & 0.695 & 1.379\\  
        FluentSpeech & 5.902 & 0.801 &  1.953 & 4.573 & 0.791 & 1.869\\
        FluentEditor & 5.896 & 0.805 &  1.969 & 4.574 & 0.794 & 1.876\\ \hline
         \textbf{FluentEditor2 }  & \textbf{5.836} & \textbf{0.815} & \textbf{1.977} & \textbf{4.537} & \textbf{0.793} & \textbf{1.875}\\  

        \bottomrule
    \end{tabular}
\vspace{-2mm}
        % 恢复行高
    \renewcommand\arraystretch{1}
    % \vspace{-6mm}
    \label{tab1}
\end{table*}

\begin{table*}[h]
    \centering
    \setstretch{0.90}
    \caption{Subjective evaluation results of comparative study.
    % * means the value achieves suboptimal.
    }\vspace{-2mm}
    \renewcommand\arraystretch{1.1}
    
    \begin{tabular}{p{2.5cm}<{\centering}|p{2.1cm}<{\centering}p{2.1cm}<{\centering}p{2.1cm}<{\centering}|p{2.1cm}<{\centering}p{2.1cm}<{\centering}p{2.1cm}<{\centering}}
        \toprule
        \multirow{2}{*}{\textbf{Method}} & \multicolumn{3}{c|}{\textbf{VCTK}} &  \multicolumn{3}{c}{\textbf{LibriTTS}} \\
         & \textbf{Insertion} & \textbf{Replacement} & \textbf{Deletion} & \textbf{Insertion} & \textbf{Replacement} & \textbf{Deletion} \\
        \midrule
        Ground Truth & 4.428 $\pm$ 0.078 & 4.422 $\pm$ 0.075 & 4.424 $\pm$ 0.078 & 4.441 $\pm$ 0.071 & 4.412 $\pm$ 0.085 & 4.417 $\pm$ 0.073 \\ \hline
        EditSpeech & 3.583 $\pm$ 0.160 & 3.634 $\pm$ 0.133 & 3.637 $\pm$ 0.177 & 3.575 $\pm$ 0.150 & 3.633 $\pm$ 0.118 & 3.542 $\pm$ 0.145 \\
        CampNet & 3.678 $\pm$ 0.120 & 3.716 $\pm$ 0.103 & 3.765 $\pm$ 0.114 & 3.601 $\pm$ 0.135 & 3.738 $\pm$ 0.117 & 3.713 $\pm$ 0.095 \\
        A$^3$T & 3.833 $\pm$ 0.104 & 3.828 $\pm$ 0.094 & 3.772 $\pm$ 0.091 & 3.597 $\pm$ 0.144 & 3.779 $\pm$ 0.110 & 3.732 $\pm$ 0.089 \\  
        FluentSpeech & 3.891 $\pm$ 0.200 & 3,936 $\pm$ 0.072 & 3.832 $\pm$ 0.157 & 3.888 $\pm$ 0.164 & 4.010 $\pm$ 0.180 & 3.983 $\pm$ 0.165 \\
        FluentEditor & 4.073 $\pm$ 0.140 & 4.114 $\pm$ 0.080 & 3.991 $\pm$ 0.139 & 4.053 $\pm$ 0.123 & 4.105 $\pm$ 0.139 & 4.044 $\pm$ 0.148  \\ \hline
        \textbf{FluentEditor2 }  & \textbf{4.314 $\pm$ 0.107} & \textbf{4.286 $\pm$ 0.088} & \textbf{4.177 $\pm$ 0.137} & \textbf{4.151 $\pm$ 0.119} & \textbf{4.191 $\pm$ 0.128} & \textbf{4.108 $\pm$ 0.124} \\  
        \bottomrule
    \end{tabular}
\vspace{-2mm}
    % 恢复行高
    \renewcommand\arraystretch{1}
    \label{tab2}
\end{table*}

For the test data, we follow \cite{chen2024diffeditor} and adopt ChatGPT \footnote{ChatGPT, based on the GPT-4 model, is accessed through OpenAI's API (GPT-4-turbo variant) for text modification tasks. For more details on the model and API, visit \url{https://openai.com}.} to modify the sentence to generate realistic and diverse $<$original sentence, modified sentence$>$ paired data, allowing for a comprehensive evaluation of the model’s ability to handle different editing sceneries.  
Please note that each system’s test set remains identical to ensure that metrics are relevant only to the speech within the editing regions, thereby maintaining accuracy and fairness in the comparative experiments.

\subsection{Comparative Study}

To comprehensively evaluate the performance of our proposed FluentEditor2, we conducted a comparative study involving five state-of-the-art neural TSE systems \footnote{We note that there was another work \cite{chen2024diffeditor} focused on TSE, but it was a contemporaneous work and hadn't been peer-reviewed and open-sourced, so we didn't add it to the baseline.}:

1) \textbf{EditSpeech} \cite{tan2021editspeech}: This system utilizes a transformer-based model for text-based speech editing. EditSpeech directly predicts the masked segments of the mel-spectrogram by leveraging the transformer’s powerful attention mechanism to reconstruct speech conditioned on both text and surrounding context.

2) \textbf{CampNet} \cite{wang2022campnet}: CampNet introduces a context-aware mask prediction network that simulates the process of speech editing by predicting the content of masked segments based on their surrounding context. The model focuses on ensuring that the predicted segments are consistent with the rest of the utterance both acoustically and contextually.

3) \textbf{A$^3$T} \cite{bai20223}: This system proposes an alignment-aware acoustic-text pre-training approach that integrates both phoneme-level information and partially-masked spectrograms as input. By combining these two modalities, A$^3$T enhances the model’s capability to accurately predict missing segments with phoneme-to-speech alignment.

4) \textbf{FluentSpeech} \cite{jiang2023fluentspeech}: FluentSpeech adopts a diffusion model framework, using a stepwise denoising process to predict the masked speech segments. By leveraging the surrounding speech as context, FluentSpeech achieves smoother and more natural-sounding transitions in edited speech.

5) \textbf{FluentEditor} \cite{liu2023fluenteditor}: FluentEditor incorporates Concatenation Loss, which combines both acoustic and prosodic consistency into the loss function, significantly improving the naturalness and coherence of the generated speech at editing boundaries.

6) \textbf{FluentEditor2 (Ours)}: Our proposed FluentEditor2 builds upon FluentEditor by introducing two new loss functions: the Hierarchical Local Acoustic Smoothness Consistency Loss ($\mathcal{L}_{HLAC}$) and the Contrastive Global Prosody Consistency Loss ($\mathcal{L}_{CGPC}$). These losses are specifically designed to ensure that edited segments integrate smoothly at both the acoustic and prosodic levels, resulting in high-quality, fluent speech.

% \begin{table*}[t]
%     \centering
%     \caption{Comparison of WER and CER Across Different Methods on VCTK and LibriTTS Datasets. \hl{add more baselines!}}
%     \begin{tabular}{p{4.8cm}<{\centering}|p{2.8cm}<{\centering}p{2.8cm}<{\centering}|p{2.8cm}<{\centering}p{2.8cm}<{\centering}}
%         \toprule
%         % \textbf{Method} & \textbf{C-FMOS}  & \textbf{MCD} $(\downarrow)$  \\
%         \multirow{2}{*}{\textbf{Method}} & \multicolumn{2}{c|}{\textbf{VCTK}} &  \multicolumn{2}{c}{\textbf{LibriTTS}} \\
%          & \textbf{WER}$(\downarrow)$  & \textbf{CER}$(\downarrow)$  & \textbf{WER}$(\downarrow)$  & \textbf{CER} $(\downarrow)$  \\
%         \midrule
%         EditSpeech  & 0.273 & 0.061 & 0.273 & 0.091\\
%         CampNet  & 0.251 & 0.128 & 0.158 & 0.072\\
%         A$^3$T  & 0.148 & 0.079 & 0.065 & 0.034\\
%         FluentSpeech  & 0.459 & 0.254 & 0.203 & 0.111\\
%         FluentEditor & 0.443 & 0.240 & 0.201 & 0.107\\ \hline
%         FluentEditor2 & \textbf{0.149} & \textbf{0.067} &  \textbf{0.029} & \textbf{0.014}\\
%         \bottomrule
%     \end{tabular}
%     % \vspace{-4mm}
%     \label{wer_cer}
% \end{table*}

To further clarify the contribution of our model, we also include the FluentEditor2-generated speech to serve as an upper bound for naturalness and fluency. To this end, five ablation systems are developed: 1) \textbf{w/o $\boldsymbol{\mathcal{L}_{HLAC}}$}, where the Hierarchical Local Acoustic Smoothness Consistency Loss is excluded, 2) \textbf{w/o $\boldsymbol{\mathcal{L}_{HLAC}}$-frame}, where the frame-level loss term of the Hierarchical Local Acoustic Smoothness Consistency Loss is excluded, 3) \textbf{w/o $\boldsymbol{\mathcal{L}_{HLAC}}$-phoneme}, where the phoneme-level loss term of the Hierarchical Local Acoustic Smoothness Consistency Loss is excluded, 4) \textbf{w/o $\boldsymbol{\mathcal{L}_{HLAC}}$-word}, where the word-level loss term of the Hierarchical Local Acoustic Smoothness Consistency Loss is excluded, and 5) \textbf{w/o $\boldsymbol{\mathcal{L}_{CGPC}}$}, where the Contrastive Global Prosody Consistency Loss is removed.
These ablation studies help us assess the individual contributions of the proposed loss functions in enhancing the fluency and naturalness of edited speech.

\iffalse 
% 修正： 展开介绍下
We develop four neural TSE systems for a comparative study, which includes: 
1) \textbf{EditSpeech} \cite{tan2021editspeech} utilizes a transformer-based model to perform text-based speech editing by directly predicting the masked segments in the spectrogram; 
2) \textbf{CampNet} \cite{wang2022campnet} propose a context-aware mask prediction network to simulate the process of text-based speech editing; 
3) A$^3$T \cite{bai20223} propose the alignment-aware acoustic-text pre-training that takes both phonemes and partially-masked spectrograms as inputs; 
4) \textbf{FluentSpeech} \cite{jiang2023fluentspeech} takes the diffusion model as backbone and predict the masked feature with the help of context speech; 
5) \textbf{FluentEditor} \cite{liu2023fluenteditor} introduces the concept of Concatenation loss and combines acoustic and prosodic consistency into the loss function to improve the naturalness and coherence of the edited speech; 
and 6) \textbf{FluentEditor2 (Ours)} designs the Local Hierarchical Acoustic Smoothness and Contrastive Global Prosody Consistency Losses. We also add the \textbf{Ground Truth} speech for comparison. Note that two ablation systems, that are ``$\boldsymbol{w/o}$ $\boldsymbol{\mathcal{L}_{HLAC}}$'' and ``$\boldsymbol{w/o}$ $\boldsymbol{\mathcal{L}_{CGPC}}$'', are built to validate the two new losses.
\fi

\section{Results and Discussion}
\label{sec:RD}

 \subsection{Evaluation of Reconstructed Speech}\label{sec:V_A}

We randomly selected 400 test samples from both the VCTK and LibriTTS datasets and reported the objective results in Table \ref{tab1}. Following the methodology of \cite{jiang2023fluentspeech}, we evaluated objective metrics for the masked regions using reconstructed speech. FluentEditor2 consistently outperformed all baselines in terms of both speech quality and fluency, achieving superior results across all three metrics: MCD, STOI, and PESQ.

For instance, FluentEditor2 achieves the lowest MCD values, indicating superior speech quality and reduced distortion, while also attaining the highest STOI and PESQ scores, reflecting better speech intelligibility and overall perceptual quality. These results highlight FluentEditor2's effectiveness in producing fluent and high-quality speech.
Note that objective metrics do not fully reflect the human perception \cite{ren2020fastspeech}, we further conduct subjective listening experiments.

\subsection{Evaluation of Edited Speech}
For the FMOS evaluation, we selected 50 audio samples for each operation (insertion, replacement, and deletion) from the test set of each dataset and invited 20 listeners to assess speech fluency. Following the methodology outlined in \cite{tan2021editspeech}, we evaluated insertion, replacement, and deletion operations, and present the FMOS results in Table \ref{tab2}. FluentEditor2 demonstrates notable improvements in fluency-related perceptual scores across both the VCTK and LibriTTS datasets. While the scores of FluentEditor2 are slightly below the ground truth values, it consistently achieves competitive results with FMOS scores of 4.314 for insertion and 4.286 for replacement, as seen in the VCTK dataset. These scores reflect the effectiveness of the fluency-aware training criteria in improving acoustic quality and prosody.

It is worth noting that the lower scores may be due to the inability to compare the edited speech with the pre-edit version directly. The perceived fluency might be even higher if comparisons could be made with the actual edited speech. Despite this, the significant advancements in FluentEditor2's fluency scores highlight the effectiveness of our approach and suggest substantial potential for further improvement.

\begin{table*}[t]
    \centering
    \caption{Objective and subjective results of ablation study.}
    % \vspace{-2mm}
    \renewcommand{\arraystretch}{0.95} % Adjust the row height here
    \begin{tabular}{p{5.3cm}|p{1.9cm}<{\centering}p{1.9cm}<{\centering}|p{1.9cm}<{\centering}p{1.9cm}<{\centering}}
        \toprule
        \multirow{2}{*}{\textbf{Method}} & \multicolumn{2}{c|}{\textbf{VCTK}} &  \multicolumn{2}{c}{\textbf{LibriTTS}} \\
         & \textbf{C-FMOS} & \textbf{MCD}$(\downarrow)$ & \textbf{C-FMOS} & \textbf{MCD}$(\downarrow)$ \\
        \midrule
        FluentEditor2 & \textbf{0.00} & \textbf{5.836} & \textbf{0.00} & \textbf{4.537}\\ \hline
        \quad w/o $ \mathcal{L}_{HLAC}$ & -0.281 & 5.917 & -0.231 & 4.544\\
    
       \quad w/o $ \mathcal{L}_{HLAC}$-frame & -0.139 & 5.900 & -0.177 & 4.594\\

    \quad w/o $ \mathcal{L}_{HLAC}$-phoneme & -0.125 & 6.004 & -0.203 & 4.564\\

    \quad w/o $ \mathcal{L}_{HLAC}$-word & -0.172 & 5.910 & -0.139 & 4.635\\ \hline
    
          \quad  w/o $ \mathcal{L}_{CGPC}$ & -0.201 & 5.896 & -0.226 & 4.587\\
        \bottomrule
    \end{tabular}
    % \vspace{-2mm}
    \label{ablation_study}
\end{table*}

\subsection{Ablation Study}\label{sec:V_C}

For the ablation study, we combined subjective and objective evaluations to assess the impact of each module. Specifically, 400 test samples were selected from the VCTK and LibriTTS datasets, with the objective metric MCD used to measure spectral differences and assess overall speech quality. For subjective evaluation, 50 edited audio inserts were randomly chosen, and C-FMOS was used to evaluate fluency and naturalness from a human perspective.

Table~\ref{ablation_study} presents results after individually removing each level of $\mathcal{L}_{\text{LHAC}}$, include frame, phoneme, and word, as well as the complete removal of $\mathcal{L}_{\text{LHAC}}$.
The results show that removing the phoneme-level constraint leads to the largest increase in MCD, reaching 6.004 for VCTK and 4.564 for LibriTTS, which underscores its crucial role in maintaining phonetic coherence and ensuring natural transitions. This highlights the importance of phoneme-level constraints in defining precise editing boundaries.
Complete removal of $\mathcal{L}_{\text{LHAC}}$ led in a further increase in MCD and a decrease in C-FMOS ($-0.281$ for VCTK and $-0.231$ for LibriTTS), indicating that the combined contributions of frame, phoneme, and word-level constraints are necessary for producing fluent, coherent speech edits. Each hierarchical level provides a unique function, with frame-level maintaining temporal alignment, phoneme-level ensuring phonetic accuracy, and word-level preserving semantic consistency.

Furthermore, the removal of $\mathcal{L}_{\text{GCPC}}$ led to a decrease in C-FMOS ($-0.201$ for VCTK and $-0.226$ for LibriTTS) and an increase in MCD, confirming its critical role in preserving prosodic quality. This constraint is essential for maintaining rhythm, intonation, and continuity, which are crucial for generating natural-sounding edits with consistent speaking style.

In summary, these findings demonstrate that both $\mathcal{L}_{\text{LHAC}}$ and $\mathcal{L}_{\text{GCPC}}$ are indispensable for ensuring the high-quality, fluent, and natural-sounding edits produced by FluentEditor2. Their combined influence allows FluentEditor2 to achieve performance that approaches the Ground Truth, significantly outperforming existing baselines.

% To further validate the contributions of our $\mathcal{L}_{\text{LHAC}}$ and $\mathcal{L}_{\text{GCPC}}$, an ablation study was conducted focusing exclusively on the insertion operation, as it consistently demonstrated superior fluency compared to replacement and deletion tasks in the subjective evaluation results. As detailed in Table \ref{ablation_study}, we report the subjective and objective outcomes after removing either $\mathcal{L}_{\text{LHAC}}$ or $\mathcal{L}_{\text{GCPC}}$ from the model.
% The results show that eliminating these constraints led to noticeable drops in both C-FMOS and MCD scores, with particularly pronounced degradations in fluency when $\mathcal{L}_{\text{LHAC}}$ was removed. This highlights the pivotal role of $\mathcal{L}_{\text{LHAC}}$ in maintaining the smoothness of speech editing, as its absence results in a substantial reduction in editing fluency. Likewise, the removal of $\mathcal{L}_{\text{GCPC}}$ underscores its importance in preserving acoustic consistency. These findings underline the critical influence of both modules in optimizing the fluency and naturalness of edited speech, confirming that their presence is essential for high-quality and fluently speech editing performance.

\begin{figure*}[thbp]
\centering
\setlength{\abovecaptionskip}{-0mm}   %调整图片标题与图距离
\begin{minipage}{0.95\linewidth}
  \centerline{
  \includegraphics[width= \linewidth]{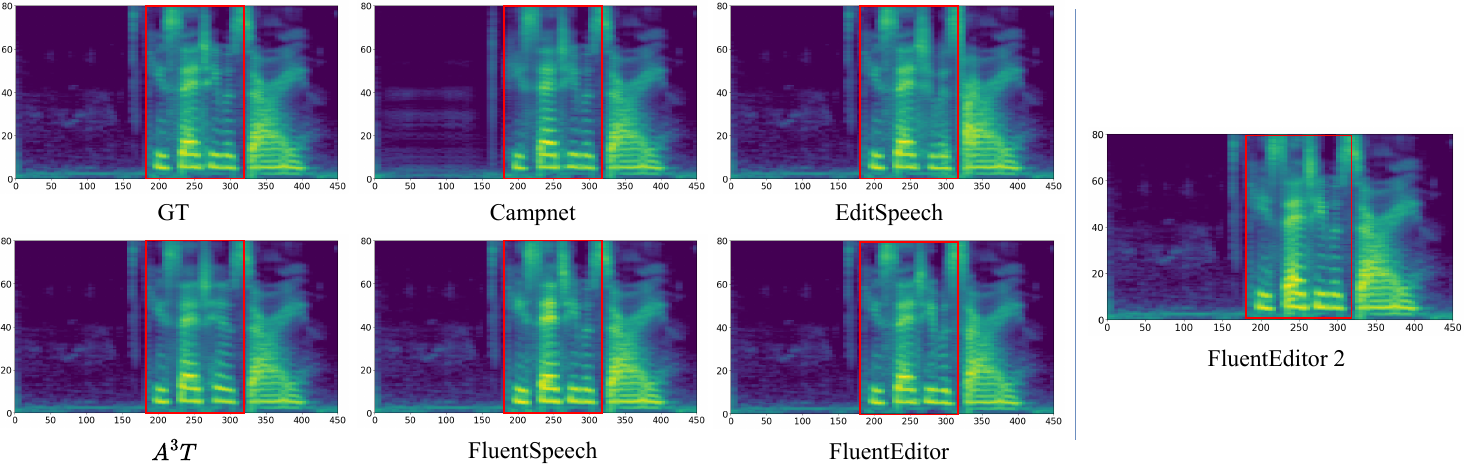}
  }
  %\centerline{(1) {\small Insert} }
\end{minipage}
\vspace{-1mm}
\caption{
Reconstruction performance comparison of mel-spectrograms generated by FluentEditor2 and baseline models.  The example shows the sentence ``he said he was sorry,'' with the red box indicating the masked and reconstructed segment ``he said he was.''
% Comparison of Reconstruction Performance for Mel-Spectrograms Between FluentEditor2 and Other Baseline Models
% The red box indicates the masked and reconstructed speech segment.
}\vspace{-2mm}
\label{fig:spec}
% \vspace       
\end{figure*}

\begin{table*}[t]
    \centering
    \caption{Evaluation of the Word-Level Masking Strategy on Baseline Models Using MCD and FMOS Metrics. } 
    \renewcommand{\arraystretch}{0.8} % Adjust the row height here
    \vspace{-2mm}
    \begin{tabular}{l|cc|cc}
        \toprule
        \multirow{2}{*}{\textbf{Method}} & \multicolumn{2}{c|}{\textbf{VCTK}} &  \multicolumn{2}{c}{\textbf{LibriTTS}} \\
         & \textbf{FMOS}  & \textbf{MCD}$(\downarrow)$  & \textbf{FMOS}  & \textbf{MCD}$(\downarrow)$ \\
        \midrule
        EditSpeech & 0.00 & 7.019 &  0.00 & 5.559 \\
        EditSpeech + WLM  & +0.029 & 7.004 &  +0.042 & 5.547 \\
        \midrule
        CampNet & 0.00 & 8.693 & 0.00 & 7.736 \\
        CampNet + WLM & +0.050 & 8.588 &  +0.005 & 7.164 \\
        \midrule
        A$^3$T & 0.00 & 6.344 &  0.00 & 5.447 \\
        A$^3$T + WLM & +0.065 & 6.192 & +0.071 & 5.306 \\
        \midrule
        FluentSpeech  & 0.00 & 5.902 & 0.00 & 4.573 \\
        FluentSpeech + WLM & +0.060 & 5.892 & +0.080 & 5.405 \\
         \midrule
        FluentEditor  & 0.00 & 5.896 &  0.00 & 4.574 \\
        FluentEditor + WLM  & +0.080 & 5.869 &  +0.075 & 4.572 \\
        % \midrule
        % FluentEditor2 & 4.087 $\pm$ 0.097 & 5.836 & 4.024 $\pm$ 0.078 & 4.537\\
        \bottomrule
    \end{tabular}
    \label{WLM}
\end{table*}

\vspace{-2mm}
\subsection{Evaluation on the Word-Level Masking Strategy.}
 % \subsection{Visualization Analysis}

% \hl{add the Word-Level Masking Strategy for all baselines. }

% A3T vs. A3T+wordMask

In this section, we attempt to demonstrate the effectiveness of Word-Level Masking strategy by also arming the WLM policy into all baselines, including A3T, EditSpeech, CampNet, FluentSpeech, and FluentEditor. We use MFA-derived duration alignment information \cite{mcauliffe2017montreal}, described in section \ref{setup}, to replace the random frame level masking strategies in the baseline with our WLM strategy.
Following the methodology used in the in the  section~\ref{sec:V_C}, the evaluation is conducted using both the objective metric MCD and the subjective metric FMOS for a comprehensive assessment. All results are reported at Table~\ref{WLM}.

% illustrates the effect of the Word-Level Masking strategy, introduced in FluentEditor2, on several baseline models, including A3T, EditSpeech, CampNet, FluentSpeech, and FluentEditor.
% \hl{The WLM skill was integrated into each baseline by replacing their original masking strategies, such as random or frame-level masking, with word-level masking to ensure alignment with word boundaries, thereby enhancing fluency and accuracy.}
% This strategy effectively improves speech quality and fluency by preserving phoneme and word boundaries, with particularly notable gains observed in FluentEditor and FluentEditor2.

In FluentEditor2, WLM enhances fluency at word boundaries by enabling more accurate alignment and splicing of word, phoneme, and frame boundaries in the edited regions. This improvement results in seamless synthesis across editing points, achieving the lowest MCD and the highest FMOS scores among all models, thereby underscoring the fluency of the generated speech. As indicated by the final row in Table ~\ref{tab1} (objective evaluation) and Table ~\ref{tab2} (subjective evaluation), FluentEditor2 exhibits superior performance, with notable reductions in distortion and significant improvements in fluency compared to other models.

While FluentEditor2 benefits most from WLM, the strategy also improves other models.  In FluentEditor, WLM reduces MCD from 5.896 to 5.869 on the VCTK dataset and from 4.574 to 4.572 on the LibriTTS dataset, facilitating more fluid word transitions and higher FMOS scores.  This reduction in MCD suggests that WLM decreases spectral distortion and improves speech fluency.  Notably, FluentEditor transitions from phoneme-level to word-level masking, highlighting the importance of masking entire words for more coherent and fluent speech synthesis.
When applied to other models like A3T, EditSpeech, CampNet, and FluentSpeech, WLM consistently improves both MCD and FMOS. These models, which originally employed non-continuous frame-level masking strategies, often disrupted phoneme and word boundaries. The introduction of WLM significantly amplifies the benefits by preserving the integrity of these boundaries, thus enhancing speech synthesis performance.

\begin{table}[t]
    \centering
    \caption{Performance Comparison at Different Mask Ratios Based on MCD, STOI, and PESQ Metrics.}
    \renewcommand{\arraystretch}{0.6} % Adjust the row height here
    \vspace{-2mm}
    \begin{tabular}{p{2.1cm}<{\centering}|p{1.6cm}<{\centering}p{1.6cm}<{\centering}p{1.6cm}<{\centering}}
        \toprule
        % \textbf{Method} & \textbf{C-FMOS}  & \textbf{MCD} $(\downarrow)$  \\
        {\textbf{Mask Ratio}} & \textbf{MCD} $(\downarrow)$   & \textbf{STOI} $(\uparrow)$  & \textbf{PESQ} $(\uparrow)$  \\
        \midrule
        % EditSpeech  & 0.472 & 0.274 & -0.226 & 4.587\\
        % CampNet  & 0.472 & 0.274 & 0.100 & 0.039\\
        % A$^3$T  & 0.472 & 0.274 & 0.089 & 0.031\\
        60\% & 5.968 & 0.805 & 1.913\\
        65\% & 5.934 & 0.807 & 1.937\\
        70\% & 5.887 & 0.810 & 1.971\\
        75\%& 5.970 & 0.808 & 1.931\\
        \textbf{80\%} & \textbf{5.836} & \textbf{0.815} & \textbf{1.977} \\
        85\%& 5.947 & 0.808 & 1.953\\
        90\%& 5.966 & 0.812 & 1.956\\
        95\%& 5.987 & 0.806 & 1.919\\
        \bottomrule
    \end{tabular}
\vspace{-2mm}
    % \vspace{-4mm}
    \label{mask}
\end{table}

\vspace{-2mm}

\subsection{Effectiveness of Different Mask Ratios in Word-Level Masking}

Although \cite{bai20223} reported that 80\% was the optimal masking rate for frame-level random selection, we extended our experiments to test similar rates for word-level masking. We varied the masking rate upwards to 60\% and downwards to 95\% to evaluate the impact of different masking ratios on speech quality and intelligibility.
For the effectiveness of different mask ratios in word-level masking, the test data was prepared in the same way as in the  section~\ref{sec:V_A}.

As shown in Table \ref{mask}, our results confirm that an 80\% mask ratio yields the best overall performance, achieving the lowest MCD (5.836) and the highest STOI (0.815) and PESQ (1.977) scores. This suggests that similar to frame-level masking, word-level masking at 80\% offers a strong balance between maintaining speech intelligibility and fluency.
When masking ratios deviate from 80\%, the performance declines slightly. Lower mask ratios (e.g., 60\%–75\%) show a higher MCD and lower PESQ, indicating less effective masking, which fails to maintain fluency adequately. Similarly, higher mask ratios (e.g., 90\%–95\%) also lead to increased MCD and reduced fluency, likely due to excessive masking disrupting the speech's natural rhythm. These results highlight the importance of selecting an optimal mask ratio, as both under- and over-masking can negatively impact the overall speech quality.
In conclusion, the experiments confirm that an 80\% mask ratio is the most effective for word-level masking, reinforcing findings from frame-level masking studies and validating the utility of this ratio across different linguistic levels.

\begin{figure*}[th!]
\centering
\setlength{\abovecaptionskip}{0mm}   %调整图片标题与图距离
\begin{minipage}{0.99\linewidth}
  \centerline{
  \includegraphics[width=\linewidth]{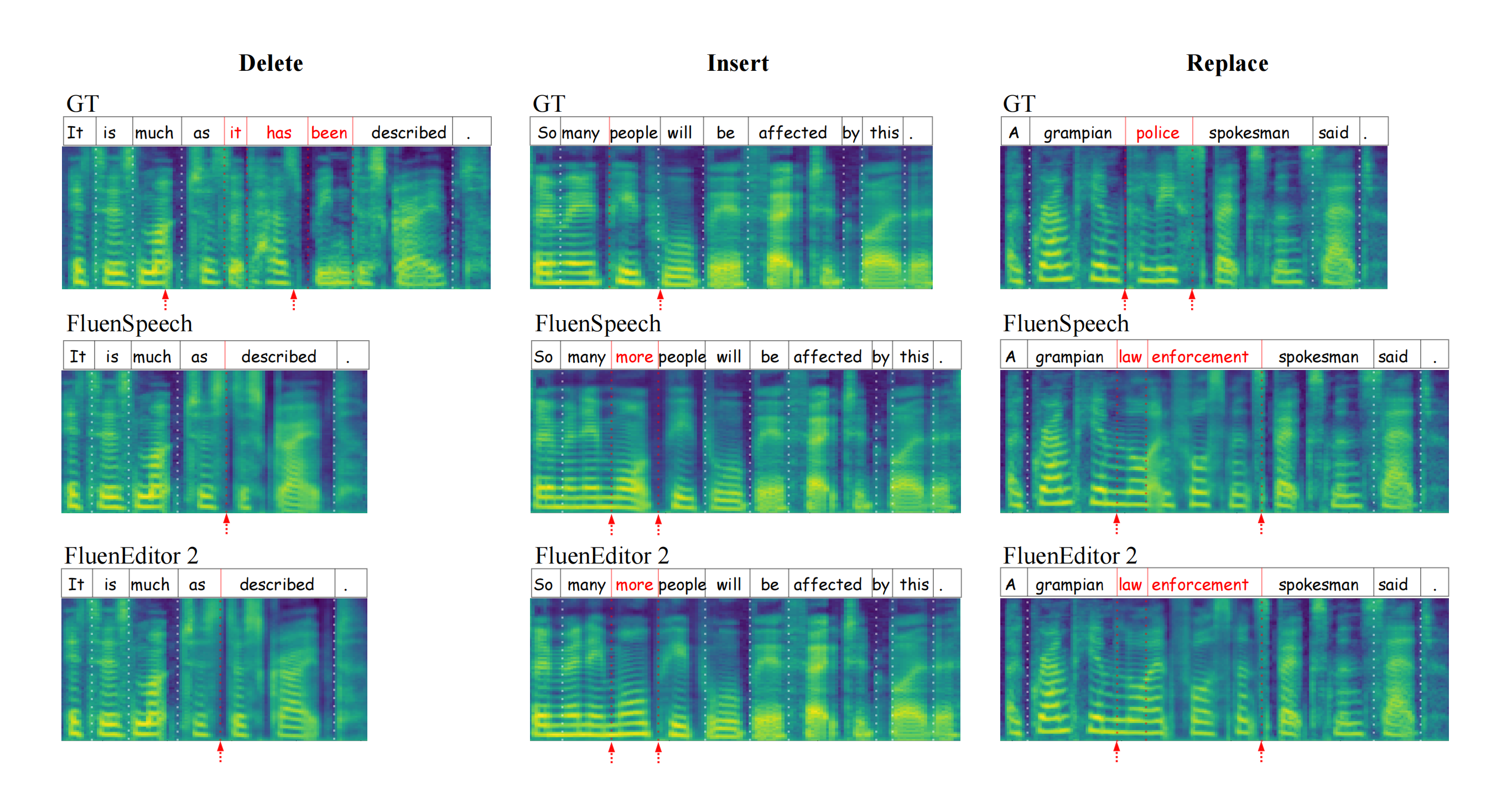}
  }
  %\centerline{(1) {\small Insert} }
\end{minipage}
\vspace{-4mm}
\caption{Visualization of editing effects on mel-spectrograms for speech insertion, replacement, and deletion. Dotted lines represent time step divisions aligned with each word in the sentence. Red highlights indicate the boundaries of the edited regions where the operations were applied.}
\vspace{-4mm}
\label{fig:edited}
% \vspace       
\end{figure*}

\subsection{Visualization Analysis of Mel-Spectrogram Reconstruction and Editing}
% \subsection{Visualization Analysis}

% {\color{blue} 

For the objective evaluation of mel-spectrograms, we randomly selected one audio sample from the 400 test samples we previously selected from both the VCTK and LibriTTS datasets (as reported in Table I). This sample was then visualized to illustrate the reconstruction performance. 

As illustrated in Fig. \ref{fig:spec}, comparing mel-spectrograms generated by FluentEditor2  and other baseline models demonstrates the superior reconstruction performance of FluentEditor2. Specifically, FluentEditor2 and consistently captures more intricate details in the frequency domain. This results in mel-spectrograms that exhibit richer spectral characteristics, leading to more natural-sounding speech outputs. For example, baseline models like EditSpeech and A$^3$T show relatively coarse spectral reconstructions, which may result in artifacts or unnatural prosody during playback. In contrast, FluentEditor2 maintains both acoustic and prosodic consistency by using acoustic and prosody loss functions, effectively capturing boundary information during the speech reconstruction process. 
For instance, Fig. \ref{fig:spec} shows the sentence ``he said he was sorry,'' with red boxes highlighting the masked and reconstructed ``he said he was'' region, clearly showcasing the enhanced smoothness and fidelity of FluentEditor2 and compared to the FluentSpeech baseline.
  % The red boxes highlight regions where speech was masked and reconstructed, clearly showcasing the enhanced smoothness and fidelity of FluentEditor2 and compared to the FluentSpeech baseline.
  % “he said he was sorry”

% {\color{blue} 
To further analyze FluentEditor2's editing capabilities and provide a clearer view of the smooth transitions in the edited regions, we randomly selected one audio sample from each operation (insertion, deletion, and replacement) from the test set of the VCTK dataset for visualization. Fig. \ref{fig:edited} provides a detailed visualization of the editing capabilities of FluentEditor2  on mel-spectrograms for common operations like deletion, insertion, and replacement of words or phrases\footnote{Deletion operations result in a reduction in spectrogram length, as segments are removed, whereas insertion leads to a lengthened spectrogram due to added content.
For replacement operations, the spectrogram length varies based on the relative duration of the new versus replaced content. Our model, Fluenteditor2, shows smoother transitions at the editing boundaries compared to baselines, with more continuous spectral patterns and fewer abrupt changes.}. The red boxes indicate the edited segments in the speech. When comparing the FluentSpeech baseline with FluentEditor2, it becomes evident that the latter can handle these modifications much more gracefully. In particular:
  \begin{itemize}
      \item For \textbf{deletion}, FluentEditor2 and smoothly removes the masked segments, ensuring that the surrounding speech remains coherent and maintains the natural prosody, unlike the baseline, where distortions can sometimes occur at the boundary.
      \item During \textbf{insertion}, the transitions between the existing and newly inserted speech segments in FluentEditor2 and are nearly seamless, preserving both prosodic rhythm and intonation. In contrast, the baseline struggles with smooth integration, leading to noticeable disruptions in speech flow.
      \item For \textbf{replacement}, FluentEditor2 and delivers a well-integrated substitution of words or phrases, resulting in consistent intonation and pacing, while the baseline may exhibit prosody mismatches or abrupt transitions.
  \end{itemize}

 These visualizations not only confirm FluentEditor2's effectiveness in reconstructing high-quality mel-spectrograms but also underscore its superior ability to handle intricate speech edits. The detailed prosodic and acoustic features retained by the model are crucial for producing natural, expressive speech even after complex modifications. More visualization results and speech samples are referred to our website: \url{https://github.com/Ai-S2-Lab/FluentEditor2}. to appreciate the advantages fully.
  
 % As illustrated in Fig.\ref{fig:spec}, we visualize the mel-spectrograms produced by FluentEditor2 and the FluentSpeech baseline. The red boxes indicate the randomly masked and reconstructed speech segment of the utterance "Scottish Women appear at Eden Court, Inverness, tonight." These masked regions are designed to capture more boundary information, allowing for a more accurate reconstruction of the speech. We can see that FluentEditor2 generates mel-spectrograms with richer frequency details compared to the baseline, resulting in more natural and expressive sounds. This demonstrates the effectiveness of our acoustic and prosody consistency losses. 

 % In FluentEditor2 and, we visualize the editing effects of the same utterance.  The red boxes highlight the segments where edits such as word or phrase insertions, replacements, or deletions have occurred. These masked regions are strategically selected to capture boundary information, enhancing the model's ability to integrate the edits into the surrounding speech seamlessly. FluentEditor2 produces smoother transitions and maintains the prosody more consistently than the baseline, which shows its superior ability to handle speech edits seamlessly. 
 % Nevertheless, we recommend that the reader listen to our speech samples\footnote{Due to space limits, we just report the FluentSpeech baseline. More visualization results and speech samples are referred to our website: \url{https://github.com/Ai-S2-Lab/FluentEditor+}. } to fully appreciate the advantages.

\begin{table*}[t]
    \centering
    \setstretch{0.88}
    \caption{Generalization Performance of FluentSpeech on Source-Domain and Cross-Domain Datasets: Comparison of MOS, FMOS, and IMOS across baseline models and FluentSpeech in both Source and Cross Domain Settings.
    % * means the value achieves suboptimal.
    }
    \renewcommand{\arraystretch}{1.1} % Adjust the row height here
   \begin{tabular}{p{2.5cm}<{\centering}|p{2.1cm}<{\centering}p{2.1cm}<{\centering}p{2.1cm}<{\centering}|p{2.1cm}<{\centering}p{2.1cm}<{\centering}p{2.1cm}<{\centering}}
        \toprule
        \multirow{2}{*}{\textbf{Method}} & \multicolumn{3}{c|}{\textbf{Source-Domain Dataset}} &  \multicolumn{3}{c}{\textbf{Cross-Domain Dataset}} \\
         & \textbf{MOS}   & \textbf{FMOS}  & \textbf{IMOS}  & \textbf{MOS}   & \textbf{FMOS}  & \textbf{IMOS}  \\
        \midrule
        % Ground Truth & NA & NA & NA & NA  & NA & NA \\ \hline
        EditSpeech & 3.718 $\pm$ 0.039 & 3.739 $\pm$ 0.039 & 3.729 $\pm$ 0.046 & 3.618 $\pm$ 0.058 & 3.630 $\pm$ 0.057 & 3.630 $\pm$ 0.060  \\
        CampNet & 3.582 $\pm$ 0.047 & 3.589 $\pm$ 0.077 & 3.574 $\pm$ 0.051 & 3.420 $\pm$ 0.139 & 3.315 $\pm$ 0.201 & 3.444 $\pm$ 0.140 \\
        A$^3$T & 3.835 $\pm$ 0.042 & 3.852 $\pm$ 0.045 & 3.862 $\pm$ 0.036 & 3.738 $\pm$ 0.065 & 3.742 $\pm$ 0.064 & 3.767 $\pm$ 0.064 \\  
        FluentSpeech & 3.876 $\pm$ 0.08 & 3.930 $\pm$ 0.084 & 3.886 $\pm$ 0.084 & 3.806 $\pm$ 0.116 & 3.800 $\pm$ 0.122 & 3.797 $\pm$ 0.123 \\
        FluentEditor & 4.052 $\pm$ 0.066 & 4.089 $\pm$ 0.075 & 4.059 $\pm$ 0.068 & 3.945 $\pm$ 0.116 &  3.974 $\pm$ 0.118 & 3.976 $\pm$ 0.124 \\ \hline
        \textbf{FluentEditor2 } & \textbf{4.247 $\pm$ 0.057} & \textbf{4.286 $\pm$ 0.066} & \textbf{4.259 $\pm$ 0.063} & \textbf{4.123 $\pm$ 0.067} & \textbf{4.135 $\pm$ 0.064} & \textbf{4.148 $\pm$ 0.068} \\  

        \bottomrule
    \end{tabular}
    \renewcommand\arraystretch{1}
    % \vspace{-6mm}
    \label{source_cross}
\end{table*}

\vspace{-2mm}
\subsection{Generalization Study}

For the generalization evaluation, we utilize LJSpeech as the cross-domain dataset \cite{ljspeech17}. Note that LJSpeech primarily contains speech related to Newgate Prison in London, which focuses differently on VCTK dataset.
For LJSpeech, we randomly select 50 samples from the test set, consisting of 20 insertions, 20 replacements, and 10 deletions. 
For VCTK, we also selected 50 samples from the test set with the same operations as the source-domain dataset.
Both source-domain and cross-domain datasets are evaluated using MOS, IMOS, and FMOS. 
Each audio sample was assessed by at least 20 listeners to evaluate speech quality and fluency. The results are presented in Table~\ref{source_cross}.

The results show that FluentEditor2 outperforms all baseline models in both settings. It achieves the highest MOS (speech quality), FMOS (fluency), and IMOS (intelligibility), demonstrating significant improvements in both perceptual and acoustic consistency. However, Cross-Domain performance shows a noticeable drop across all models, particularly in MOS and FMOS, highlighting the limited ability of existing methods to generalize effectively to Cross-Domain scenarios. This performance gap can be attributed to the fact that most models are primarily trained and evaluated within the same domain, limiting their robustness when faced with unseen data from a different domain.

\section{Conclusion}
\label{sec:con}
 In this paper, we introduced FluentEditor2, a novel text-based speech editing (TSE) model designed to improve the acoustic and prosody consistency of edited speech through two innovative fluency-aware training criteria. The proposed Hierarchical Local Acoustic Smoothness Consistency Loss ($\mathcal{L}_{HLAC}$) evaluates the consistency of acoustic features at editing boundaries both at the frame and phoneme levels, ensuring that the transitions at concatenation points are smooth and natural. Additionally, the Contrastive Global Prosody Consistency Loss ($\mathcal{L}_{CGPC}$) uses contrastive learning to align the high-level prosodic features within the edited regions with the surrounding context, while distinctly separating them from irrelevant segments. 
 We validated the effectiveness of FluentEditor2 through extensive objective and subjective experiments on two datasets, VCTK and LibriTTS, covering the three core editing operations: insertion, replacement, and deletion. The results consistently demonstrated that incorporating $\mathcal{L}_{HLAC}$ and $\mathcal{L}_{CGPC}$ significantly enhances the fluency and prosody of edited speech across all datasets and operations. This comprehensive evaluation shows that FluentEditor2 not only produces more natural and seamless transitions but also maintains consistent prosody, providing a superior solution for text-based speech editing.

% In this paper, we introduce a novel text-based speech editing (TSE) model, termed FluentEditor2, that involves two novel fluency-aware training criteria to improve the acoustic and prosody consistency of edited speech. 
% The Hierarchical Local Acoustic Smoothness Consistency Loss $\mathcal{L}_{HLAC}$ assesses the acoustic features of editing boundaries hierarchically, at both the frame and phoneme levels, to determine their consistency with the acoustic features at the actual concatenation points. On the other hand, the Contrastive Global Prosody Consistency Loss $\mathcal{L}_{CGPC}$ aims to ensure through contrastive learning that the high-level prosodic features of the synthesized audio within the editing region closely resemble those of the original context, while diverging from the prosodic features of irrelevant audio segments.

% We conduct objective and subjective experiments on two datasets, VCTK and LibriTTS, involving three editing operations: insertion, replacement, and deletion. The results demonstrate that incorporating $\mathcal{L}_{HLAC}$ and $\mathcal{L}_{CGPC}$ yields superior results and ensures fluent speech with consistent prosody across both datasets and all three editing operations.

% The objective and subjective experiments on VCTK demonstrate that incorporating $\mathcal{L}_{HLAC}$ and $\mathcal{L}_{CGPC}$ yields superior results and ensures fluent speech with consistent prosody. 

% In future work, we will consider the multi-scale consistency and further improve the FluentEditor architecture.

\normalem
\bibliographystyle{IEEEtran}
{\footnotesize
\bibliography{refs}}

\begin{thebibliography}{10}
\providecommand{\url}[1]{#1}
\csname url@samestyle\endcsname
\providecommand{\newblock}{\relax}
\providecommand{\bibinfo}[2]{#2}
\providecommand{\BIBentrySTDinterwordspacing}{\spaceskip=0pt\relax}
\providecommand{\BIBentryALTinterwordstretchfactor}{4}
\providecommand{\BIBentryALTinterwordspacing}{\spaceskip=\fontdimen2\font plus
\BIBentryALTinterwordstretchfactor\fontdimen3\font minus
  \fontdimen4\font\relax}
\providecommand{\BIBforeignlanguage}[2]{{%
\expandafter\ifx\csname l@#1\endcsname\relax
\typeout{** WARNING: IEEEtran.bst: No hyphenation pattern has been}%
\typeout{** loaded for the language `#1'. Using the pattern for}%
\typeout{** the default language instead.}%
\else
\language=\csname l@#1\endcsname
\fi
#2}}
\providecommand{\BIBdecl}{\relax}
\BIBdecl

\bibitem{jin2017voco}
Z.~Jin, G.~J. Mysore, S.~Diverdi, J.~Lu, and A.~Finkelstein, ``Voco: Text-based
  insertion and replacement in audio narration,'' \emph{ACM Transactions on
  Graphics (TOG)}, vol.~36, no.~4, pp. 1--13, 2017.

\bibitem{10379131}
R.~Liu, Y.~Hu, H.~Zuo, Z.~Luo, L.~Wang, and G.~Gao, ``Text-to-speech for
  low-resource agglutinative language with morphology-aware language model
  pre-training,'' \emph{IEEE/ACM Transactions on Audio, Speech, and Language
  Processing}, vol.~32, pp. 1075--1087, 2024.

\bibitem{10487819}
R.~Liu, B.~Sisman, G.~Gao, and H.~Li, ``Controllable accented text-to-speech
  synthesis with fine and coarse-grained intensity rendering,'' \emph{IEEE/ACM
  Transactions on Audio, Speech, and Language Processing}, vol.~32, pp.
  2188--2201, 2024.

\bibitem{mcauliffe2017montreal}
M.~McAuliffe, M.~Socolof, S.~Mihuc, M.~Wagner, and M.~Sonderegger, ``Montreal
  forced aligner: Trainable text-speech alignment using kaldi.'' \emph{Proc.
  Interspeech 2017}, pp. 498--502, 2017.

\bibitem{jiang2023fluentspeech}
\BIBentryALTinterwordspacing
Z.~Jiang, Q.~Yang, J.~Zuo, Z.~Ye, R.~Huang, Y.~Ren, and Z.~Zhao,
  ``{F}luent{S}peech: Stutter-oriented automatic speech editing with
  context-aware diffusion models,'' in \emph{Findings of the Association for
  Computational Linguistics: ACL 2023}.\hskip 1em plus 0.5em minus 0.4em\relax
  Toronto, Canada: Association for Computational Linguistics, Jul. 2023, pp.
  11\,655--11\,671. [Online]. Available:
  \url{https://aclanthology.org/2023.findings-acl.741}
\BIBentrySTDinterwordspacing

\bibitem{liu2023fluenteditor}
R.~Liu, J.~Xi, Z.~Jiang, and H.~Li, ``Fluenteditor: Text-based speech editing
  by considering acoustic and prosody consistency,'' in \emph{Interspeech
  2024}, 2024, pp. 3435--3439.

\bibitem{morrison2021context}
M.~Morrison, L.~Rencker, Z.~Jin, N.~J. Bryan, J.-P. Caceres, and B.~Pardo,
  ``Context-aware prosody correction for text-based speech editing,'' in
  \emph{ICASSP 2021-2021 IEEE International Conference on Acoustics, Speech and
  Signal Processing (ICASSP)}.\hskip 1em plus 0.5em minus 0.4em\relax IEEE,
  2021, pp. 7038--7042.

\bibitem{tan2021editspeech}
D.~Tan, L.~Deng, Y.~T. Yeung, X.~Jiang, X.~Chen, and T.~Lee, ``Editspeech: A
  text based speech editing system using partial inference and bidirectional
  fusion,'' in \emph{2021 IEEE Automatic Speech Recognition and Understanding
  Workshop (ASRU)}.\hskip 1em plus 0.5em minus 0.4em\relax IEEE, 2021, pp.
  626--633.

\bibitem{wang2022campnet}
T.~Wang, J.~Yi, R.~Fu, J.~Tao, and Z.~Wen, ``Campnet: Context-aware mask
  prediction for end-to-end text-based speech editing,'' \emph{IEEE/ACM
  Transactions on Audio, Speech, and Language Processing}, vol.~30, pp.
  2241--2254, 2022.

\bibitem{bai20223}
H.~Bai, R.~Zheng, J.~Chen, M.~Ma, X.~Li, and L.~Huang, ``${A}^{3}{T}$:
  Alignment-aware acoustic and text pretraining for speech synthesis and
  editing,'' in \emph{International Conference on Machine Learning}.\hskip 1em
  plus 0.5em minus 0.4em\relax PMLR, 2022, pp. 1399--1411.

\bibitem{tae22_interspeech}
J.~Tae, H.~Kim, and T.~Kim, ``Editts: Score-based editing for controllable
  text-to-speech,'' in \emph{Interspeech 2022}, 2022, pp. 421--425.

\bibitem{zen19_interspeech}
H.~Zen, V.~Dang, R.~Clark, Y.~Zhang, R.~J. Weiss, Y.~Jia, Z.~Chen, and Y.~Wu,
  ``Libritts: A corpus derived from librispeech for text-to-speech,'' in
  \emph{Interspeech 2019}, 2019, pp. 1526--1530.

\bibitem{reyes2016spectrum}
H.~Reyes, S.~Subramaniam, N.~Kaabouch, and W.~C. Hu, ``A spectrum sensing
  technique based on autocorrelation and euclidean distance and its comparison
  with energy detection for cognitive radio networks,'' \emph{Computers \&
  Electrical Engineering}, vol.~52, pp. 319--327, 2016.

\bibitem{tseng2005fluent}
C.-y. Tseng, S.-h. Pin, Y.~Lee, H.-m. Wang, and Y.-c. Chen, ``Fluent speech
  prosody: Framework and modeling,'' \emph{Speech communication}, vol.~46, no.
  3-4, pp. 284--309, 2005.

\bibitem{liu2021expressive}
R.~Liu, B.~Sisman, G.~Gao, and H.~Li, ``Expressive tts training with frame and
  style reconstruction loss,'' \emph{IEEE/ACM Transactions on Audio, Speech,
  and Language Processing}, vol.~29, pp. 1806--1818, 2021.

\bibitem{glass1988multi}
J.~R. Glass and V.~W. Zue, ``Multi-level acoustic segmentation of continuous
  speech,'' in \emph{ICASSP-88., International Conference on Acoustics, Speech,
  and Signal Processing}.\hskip 1em plus 0.5em minus 0.4em\relax IEEE Computer
  Society, 1988, pp. 429--430.

\bibitem{rabiner2007introduction}
L.~R. Rabiner, R.~W. Schafer \emph{et~al.}, ``Introduction to digital speech
  processing,'' \emph{Foundations and Trends{\textregistered} in Signal
  Processing}, vol.~1, no. 1--2, pp. 1--194, 2007.

\bibitem{honda2004physiological}
K.~Honda, ``Physiological factors causing tonal characteristics of speech: from
  global to local prosody,'' in \emph{Speech Prosody 2004, International
  Conference}, 2004.

\bibitem{qian2021global}
K.~Qian, Y.~Zhang, S.~Chang, J.~Xiong, C.~Gan, D.~Cox, and M.~Hasegawa-Johnson,
  ``Global prosody style transfer without text transcriptions,'' in
  \emph{International Conference on Machine Learning}.\hskip 1em plus 0.5em
  minus 0.4em\relax PMLR, 2021, pp. 8650--8660.

\bibitem{blouin2002concatenation}
C.~Blouin, O.~Rosec, P.~C. Bagshaw, and C.~d’Alessandro, ``Concatenation cost
  calculation and optimisation for unit selection in tts,'' in \emph{IEEE
  workshop on speech synthesis}, 2002, pp. 0--3.

\bibitem{hunt1996unit}
A.~J. Hunt and A.~W. Black, ``Unit selection in a concatenative speech
  synthesis system using a large speech database,'' in \emph{1996 IEEE
  international conference on acoustics, speech, and signal processing
  conference proceedings}, vol.~1.\hskip 1em plus 0.5em minus 0.4em\relax IEEE,
  1996, pp. 373--376.

\bibitem{wouters2000unit}
J.~Wouters and M.~W. Macon, ``Unit fusion for concatenative speech synthesis.''
  in \emph{INTERSPEECH}.\hskip 1em plus 0.5em minus 0.4em\relax Citeseer, 2000,
  pp. 302--305.

\bibitem{dong2006unit}
M.~Dong, K.-T. Lua, and H.~Li, ``A unit selection-based speech synthesis
  approach for mandarin chinese.'' \emph{J. Chin. Lang. Comput.}, vol.~16,
  no.~3, pp. 135--144, 2006.

\bibitem{fu2018deep}
R.~Fu, J.~Tao, Y.~Zheng, and Z.~Wen, ``Deep metric learning for the target cost
  in unit-selection speech synthesizer.'' in \emph{INTERSPEECH}, 2018, pp.
  2514--2518.

\bibitem{chappell2002comparison}
D.~T. Chappell and J.~H. Hansen, ``A comparison of spectral smoothing methods
  for segment concatenation based speech synthesis,'' \emph{Speech
  Communication}, vol.~36, no. 3-4, pp. 343--373, 2002.

\bibitem{le2020contrastive}
P.~H. Le-Khac, G.~Healy, and A.~F. Smeaton, ``Contrastive representation
  learning: A framework and review,'' \emph{Ieee Access}, vol.~8, pp.
  193\,907--193\,934, 2020.

\bibitem{veaux2017cstr}
C.~Veaux, J.~Yamagishi, K.~MacDonald \emph{et~al.}, ``Cstr vctk corpus: English
  multi-speaker corpus for cstr voice cloning toolkit,'' \emph{University of
  Edinburgh. The Centre for Speech Technology Research (CSTR)}, vol.~6, p.~15,
  2017.

\bibitem{chapman2016statistical}
B.~P. Chapman, A.~Weiss, and P.~R. Duberstein, ``Statistical learning theory
  for high dimensional prediction: Application to criterion-keyed scale
  development.'' \emph{Psychological methods}, vol.~21, no.~4, p. 603, 2016.

\bibitem{whittaker2004semantic}
S.~Whittaker and B.~Amento, ``Semantic speech editing,'' in \emph{Proceedings
  of the SIGCHI conference on Human factors in computing systems}, 2004, pp.
  527--534.

\bibitem{rubin2013content}
S.~Rubin, F.~Berthouzoz, G.~J. Mysore, W.~Li, and M.~Agrawala, ``Content-based
  tools for editing audio stories,'' in \emph{Proceedings of the 26th annual
  ACM symposium on User interface software and technology}, 2013, pp. 113--122.

\bibitem{baume2018contextual}
C.~Baume, M.~D. Plumbley, J.~{\'C}ali{\'c}, and D.~Frohlich, ``A contextual
  study of semantic speech editing in radio production,'' \emph{International
  Journal of Human-Computer Studies}, vol. 115, pp. 67--80, 2018.

\bibitem{descript}
``Descript,'' \url{https://www.descript.com/}, 2020.

\bibitem{moulines1990pitch}
E.~Moulines and F.~Charpentier, ``Pitch-synchronous waveform processing
  techniques for text-to-speech synthesis using diphones,'' \emph{Speech
  communication}, vol.~9, no. 5-6, pp. 453--467, 1990.

\bibitem{sisman2020overview}
B.~Sisman, J.~Yamagishi, S.~King, and H.~Li, ``An overview of voice conversion
  and its challenges: From statistical modeling to deep learning,''
  \emph{IEEE/ACM Transactions on Audio, Speech, and Language Processing},
  vol.~29, pp. 132--157, 2020.

\bibitem{mohammadi2017overview}
S.~H. Mohammadi and A.~Kain, ``An overview of voice conversion systems,''
  \emph{Speech Communication}, vol.~88, pp. 65--82, 2017.

\bibitem{sun2016phonetic}
L.~Sun, K.~Li, H.~Wang, S.~Kang, and H.~Meng, ``Phonetic posteriorgrams for
  many-to-one voice conversion without parallel data training,'' in \emph{2016
  IEEE International Conference on Multimedia and Expo (ICME)}.\hskip 1em plus
  0.5em minus 0.4em\relax IEEE, 2016, pp. 1--6.

\bibitem{jin2018speech}
Z.~Jin \emph{et~al.}, ``Speech synthesis for text-based editing of audio
  narration,'' Ph.D. dissertation, Doctoral dissertation, Ph. D. dissertation,
  Comput. Sci. Dept., Princeton~…, 2018.

\bibitem{gulati20_interspeech}
A.~Gulati, J.~Qin, C.-C. Chiu, N.~Parmar, Y.~Zhang, J.~Yu, W.~Han, S.~Wang,
  Z.~Zhang, Y.~Wu, and R.~Pang, ``Conformer: Convolution-augmented transformer
  for speech recognition,'' in \emph{Interspeech 2020}, 2020, pp. 5036--5040.

\bibitem{guo2021recent}
P.~Guo, F.~Boyer, X.~Chang, T.~Hayashi, Y.~Higuchi, H.~Inaguma, N.~Kamo, C.~Li,
  D.~Garcia-Romero, J.~Shi \emph{et~al.}, ``Recent developments on espnet
  toolkit boosted by conformer,'' in \emph{ICASSP 2021-2021 IEEE International
  Conference on Acoustics, Speech and Signal Processing (ICASSP)}.\hskip 1em
  plus 0.5em minus 0.4em\relax IEEE, 2021, pp. 5874--5878.

\bibitem{croitoru2023diffusion}
F.-A. Croitoru, V.~Hondru, R.~T. Ionescu, and M.~Shah, ``Diffusion models in
  vision: A survey,'' \emph{IEEE Transactions on Pattern Analysis and Machine
  Intelligence}, vol.~45, no.~9, pp. 10\,850--10\,869, 2023.

\bibitem{jeong21_interspeech}
M.~Jeong, H.~Kim, S.~J. Cheon, B.~J. Choi, and N.~S. Kim, ``Diff-tts: A
  denoising diffusion model for text-to-speech,'' in \emph{Interspeech 2021},
  2021, pp. 3605--3609.

\bibitem{popov2021grad}
V.~Popov, I.~Vovk, V.~Gogoryan, T.~Sadekova, and M.~Kudinov, ``Grad-tts: A
  diffusion probabilistic model for text-to-speech,'' in \emph{International
  Conference on Machine Learning}.\hskip 1em plus 0.5em minus 0.4em\relax PMLR,
  2021, pp. 8599--8608.

\bibitem{DBLP:journals/corr/abs-2403-04804}
\BIBentryALTinterwordspacing
A.~Alexos and P.~Baldi, ``Attentionstitch: How attention solves the speech
  editing problem,'' \emph{CoRR}, vol. abs/2403.04804, 2024. [Online].
  Available: \url{https://doi.org/10.48550/arXiv.2403.04804}
\BIBentrySTDinterwordspacing

\bibitem{simpson1989lexical}
G.~B. Simpson, R.~R. Peterson, M.~A. Casteel, and C.~Burgess, ``Lexical and
  sentence context effects in word recognition.'' \emph{Journal of Experimental
  Psychology: Learning, Memory, and Cognition}, vol.~15, no.~1, p.~88, 1989.

\bibitem{schafer1975digital}
R.~W. Schafer and L.~R. Rabiner, ``Digital representations of speech signals,''
  \emph{Proceedings of the IEEE}, vol.~63, no.~4, pp. 662--677, 1975.

\bibitem{lin2020automatic}
B.~Lin, L.~Wang, X.~Feng, and J.~Zhang, ``Automatic scoring at
  multi-granularity for l2 pronunciation.'' in \emph{Interspeech}, 2020, pp.
  3022--3026.

\bibitem{chen22_interspeech}
W.~Chen, X.~Xing, X.~Xu, J.~Pang, and L.~Du, ``Speechformer: A hierarchical
  efficient framework incorporating the characteristics of speech,'' in
  \emph{Interspeech 2022}, 2022, pp. 346--350.

\bibitem{lei2022msemotts}
Y.~Lei, S.~Yang, X.~Wang, and L.~Xie, ``Msemotts: Multi-scale emotion transfer,
  prediction, and control for emotional speech synthesis,'' \emph{IEEE/ACM
  Transactions on Audio, Speech, and Language Processing}, vol.~30, pp.
  853--864, 2022.

\bibitem{hono20_interspeech}
Y.~Hono, K.~Tsuboi, K.~Sawada, K.~Hashimoto, K.~Oura, Y.~Nankaku, and
  K.~Tokuda, ``Hierarchical multi-grained generative model for expressive
  speech synthesis,'' in \emph{Interspeech 2020}, 2020, pp. 3441--3445.

\bibitem{li21r_interspeech}
X.~Li, C.~Song, J.~Li, Z.~Wu, J.~Jia, and H.~Meng, ``Towards multi-scale style
  control for expressive speech synthesis,'' in \emph{Interspeech 2021}, 2021,
  pp. 4673--4677.

\bibitem{ren2021portaspeech}
Y.~Ren, J.~Liu, and Z.~Zhao, ``Portaspeech: Portable and high-quality
  generative text-to-speech,'' \emph{Advances in Neural Information Processing
  Systems}, vol.~34, pp. 13\,963--13\,974, 2021.

\bibitem{lei2023msstyletts}
S.~Lei, Y.~Zhou, L.~Chen, Z.~Wu, X.~Wu, S.~Kang, and H.~Meng, ``Msstyletts:
  Multi-scale style modeling with hierarchical context information for
  expressive speech synthesis,'' \emph{IEEE/ACM Transactions on Audio, Speech,
  and Language Processing}, 2023.

\bibitem{jiang24d_interspeech}
Y.~Jiang, T.~Li, F.~Yang, L.~Xie, M.~Meng, and Y.~Wang, ``Towards expressive
  zero-shot speech synthesis with hierarchical prosody modeling,'' in
  \emph{Interspeech 2024}, 2024, pp. 2300--2304.

\bibitem{hadsell2006dimensionality}
R.~Hadsell, S.~Chopra, and Y.~LeCun, ``Dimensionality reduction by learning an
  invariant mapping,'' in \emph{2006 IEEE computer society conference on
  computer vision and pattern recognition (CVPR'06)}, vol.~2.\hskip 1em plus
  0.5em minus 0.4em\relax IEEE, 2006, pp. 1735--1742.

\bibitem{chen2020simple}
T.~Chen, S.~Kornblith, M.~Norouzi, and G.~Hinton, ``A simple framework for
  contrastive learning of visual representations,'' in \emph{International
  conference on machine learning}.\hskip 1em plus 0.5em minus 0.4em\relax PMLR,
  2020, pp. 1597--1607.

\bibitem{viana2023multi}
T.~B. Viana, V.~L. Souza, A.~L. Oliveira, R.~M. Cruz, and R.~Sabourin, ``A
  multi-task approach for contrastive learning of handwritten signature feature
  representations,'' \emph{Expert Systems with Applications}, vol. 217, p.
  119589, 2023.

\bibitem{chen2021exploring}
X.~Chen and K.~He, ``Exploring simple siamese representation learning,'' in
  \emph{Proceedings of the IEEE/CVF conference on computer vision and pattern
  recognition}, 2021, pp. 15\,750--15\,758.

\bibitem{yang2019reducing}
Z.~Yang, Y.~Cheng, Y.~Liu, and M.~Sun, ``Reducing word omission errors in
  neural machine translation: A contrastive learning approach,'' in
  \emph{Proceedings of the 57th Annual Meeting of the Association for
  Computational Linguistics}, 2019, pp. 6191--6196.

\bibitem{bose-etal-2018-adversarial}
\BIBentryALTinterwordspacing
A.~J. Bose, H.~Ling, and Y.~Cao, ``Adversarial contrastive estimation,'' in
  \emph{Proceedings of the 56th Annual Meeting of the Association for
  Computational Linguistics (Volume 1: Long Papers)}, I.~Gurevych and Y.~Miyao,
  Eds.\hskip 1em plus 0.5em minus 0.4em\relax Melbourne, Australia: Association
  for Computational Linguistics, Jul. 2018, pp. 1021--1032. [Online].
  Available: \url{https://aclanthology.org/P18-1094}
\BIBentrySTDinterwordspacing

\bibitem{kharitonov2021data}
E.~Kharitonov, M.~Rivi{\`e}re, G.~Synnaeve, L.~Wolf, P.-E. Mazar{\'e},
  M.~Douze, and E.~Dupoux, ``Data augmenting contrastive learning of speech
  representations in the time domain,'' in \emph{2021 IEEE Spoken Language
  Technology Workshop (SLT)}.\hskip 1em plus 0.5em minus 0.4em\relax IEEE,
  2021, pp. 215--222.

\bibitem{koepke2022audio}
A.~S. Koepke, A.-M. Oncescu, J.~F. Henriques, Z.~Akata, and S.~Albanie, ``Audio
  retrieval with natural language queries: A benchmark study,'' \emph{IEEE
  Transactions on Multimedia}, vol.~25, pp. 2675--2685, 2022.

\bibitem{wu22e_interspeech}
Y.~Wu, X.~Wang, S.~Zhang, L.~He, R.~Song, and J.-Y. Nie, ``Self-supervised
  context-aware style representation for expressive speech synthesis,'' in
  \emph{Interspeech 2022}, 2022, pp. 5503--5507.

\bibitem{meng22c_interspeech}
Y.~Meng, X.~Li, Z.~Wu, T.~Li, Z.~Sun, X.~Xiao, C.~Sun, H.~Zhan, and H.~Meng,
  ``Calm: Constrastive cross-modal speaking style modeling for expressive
  text-to-speech synthesis,'' in \emph{Interspeech 2022}, 2022, pp. 5533--5537.

\bibitem{wu2024dctts}
Z.~Wu, Q.~Li, S.~Liu, and Q.~Yang, ``Dctts: Discrete diffusion model with
  contrastive learning for text-to-speech generation,'' in \emph{ICASSP
  2024-2024 IEEE International Conference on Acoustics, Speech and Signal
  Processing (ICASSP)}.\hskip 1em plus 0.5em minus 0.4em\relax IEEE, 2024, pp.
  11\,336--11\,340.

\bibitem{ren2022revisiting}
Y.~Ren, X.~Tan, T.~Qin, Z.~Zhao, and T.-Y. Liu, ``Revisiting over-smoothness in
  text to speech,'' in \emph{Proceedings of the 60th Annual Meeting of the
  Association for Computational Linguistics (Volume 1: Long Papers)}, 2022, pp.
  8197--8213.

\bibitem{mandhare2019generalized}
P.~S. Mandhare, V.~R. Borkar, and M.~R. Kumbhakarna, ``The generalized
  approximation of an arbitrary function using its mean and quadratic
  variance,'' \emph{IEEE Transactions on Circuits and Systems II: Express
  Briefs}, vol.~66, no.~12, pp. 2096--2100, 2019.

\bibitem{hashim2021mobile}
N.~N. W.~N. Hashim, M.~A.-E.~A. Ezzi, and M.~D. Wilkes, ``Mobile microphone
  robust acoustic feature identification using coefficient of variance,''
  \emph{International Journal of Speech Technology}, vol.~24, no.~4, pp.
  1089--1100, 2021.

\bibitem{wang2018style}
Y.~Wang, D.~Stanton, Y.~Zhang, R.-S. Ryan, E.~Battenberg, J.~Shor, Y.~Xiao,
  Y.~Jia, F.~Ren, and R.~A. Saurous, ``Style tokens: Unsupervised style
  modeling, control and transfer in end-to-end speech synthesis,'' in
  \emph{International conference on machine learning}.\hskip 1em plus 0.5em
  minus 0.4em\relax PMLR, 2018, pp. 5180--5189.

\bibitem{kong2020hifi}
J.~Kong, J.~Kim, and J.~Bae, ``Hifi-gan: Generative adversarial networks for
  efficient and high fidelity speech synthesis,'' \emph{Advances in Neural
  Information Processing Systems}, vol.~33, pp. 17\,022--17\,033, 2020.

\bibitem{DBLP:journals/corr/KingmaB14}
\BIBentryALTinterwordspacing
D.~P. Kingma and J.~Ba, ``Adam: {A} method for stochastic optimization,'' in
  \emph{3rd International Conference on Learning Representations, {ICLR} 2015,
  San Diego, CA, USA, May 7-9, 2015, Conference Track Proceedings}, Y.~Bengio
  and Y.~LeCun, Eds., 2015. [Online]. Available:
  \url{http://arxiv.org/abs/1412.6980}
\BIBentrySTDinterwordspacing

\bibitem{loizou2011speech}
P.~C. Loizou, ``Speech quality assessment,'' in \emph{Multimedia analysis,
  processing and communications}.\hskip 1em plus 0.5em minus 0.4em\relax
  Springer, 2011, pp. 623--654.

\bibitem{chen2024diffeditor}
Y.~Chen, Y.~Jia, S.~Zhao, Z.~Jiang, H.~Li, J.~Kang, and Y.~Qin, ``Diffeditor:
  Enhancing speech editing with semantic enrichment and acoustic consistency,''
  \emph{arXiv preprint arXiv:2409.12992}, 2024.

\bibitem{kubichek1993mel}
R.~Kubichek, ``Mel-cepstral distance measure for objective speech quality
  assessment,'' in \emph{Proceedings of IEEE pacific rim conference on
  communications computers and signal processing}, vol.~1.\hskip 1em plus 0.5em
  minus 0.4em\relax IEEE, 1993, pp. 125--128.

\bibitem{taal2010short}
C.~H. Taal, R.~C. Hendriks, R.~Heusdens, and J.~Jensen, ``A short-time
  objective intelligibility measure for time-frequency weighted noisy speech,''
  in \emph{2010 IEEE international conference on acoustics, speech and signal
  processing}.\hskip 1em plus 0.5em minus 0.4em\relax IEEE, 2010, pp.
  4214--4217.

\bibitem{rix2001perceptual}
A.~W. Rix, J.~G. Beerends, M.~P. Hollier, and A.~P. Hekstra, ``Perceptual
  evaluation of speech quality (pesq)-a new method for speech quality
  assessment of telephone networks and codecs,'' in \emph{2001 IEEE
  international conference on acoustics, speech, and signal processing.
  Proceedings (Cat. No. 01CH37221)}, vol.~2.\hskip 1em plus 0.5em minus
  0.4em\relax IEEE, 2001, pp. 749--752.

\bibitem{ren2020fastspeech}
Y.~Ren, C.~Hu, X.~Tan, T.~Qin, S.~Zhao, Z.~Zhao, and T.-Y. Liu, ``Fastspeech 2:
  Fast and high-quality end-to-end text to speech,'' in \emph{International
  Conference on Learning Representations}, 2020.

\bibitem{ljspeech17}
K.~Ito and L.~Johnson, ``The lj speech dataset,''
  \url{https://keithito.com/LJ-Speech-Dataset/}, 2017.

\end{thebibliography}

\end{document}